%% file: pnas_submission.tex
\documentclass[12pt]{article}
\usepackage{amsmath}       
\usepackage{amssymb}       
\usepackage{bm}            
\usepackage{physics}       
\usepackage{mathtools}     
\usepackage{graphicx}      
\usepackage{booktabs}      
\usepackage{caption}       
\usepackage{subcaption}    
\usepackage{comment}       
\usepackage[margin=1in]{geometry} 
\usepackage{authblk}       
\usepackage{natbib}        
\usepackage{hyperref}      
\hypersetup{hidelinks}
\usepackage{xcolor}        
\usepackage{float}         

\begin{document}

\title{Predicting Forced Responses of Probability Distributions via the Fluctuation--Dissipation Theorem and Generative Modeling}

\author[1]{Ludovico T. Giorgini\thanks{Correspondence: ludogio@mit.edu}}
\author[2]{Fabrizio Falasca}
\author[3]{Andre N. Souza}

\affil[1]{Department of Mathematics, Massachusetts Institute of Technology, Cambridge, MA 02139, USA}
\affil[2]{Courant Institute of Mathematical Sciences, New York University, New York, NY 10012, USA}
\affil[3]{Department of Earth, Atmospheric and Planetary Sciences, Massachusetts Institute of Technology, Cambridge, MA 02139, USA}

\date{\today}

\maketitle

\begin{abstract}
We present a novel and flexible data-driven framework for estimating the response of higher-order moments of nonlinear stochastic systems to small external perturbations. The classical Generalized Fluctuation–Dissipation Theorem (GFDT) links the unperturbed steady-state distribution to the system’s linear response. While standard implementations relying on Gaussian approximations can predict the mean response, they often fail to capture changes in higher-order moments. To overcome this, we combine GFDT with score-based generative modeling to estimate the system's score function directly from data. {We demonstrate the framework's versatility by employing two complementary score estimation techniques tailored to the system's characteristics: (i) a clustering-based algorithm (KGMM) for systems with low-dimensional effective dynamics, and (ii) a denoising score matching method implemented with a U-Net architecture for high-dimensional, spatially-extended systems where reduced-order modeling is not feasible.} Our method is validated on several stochastic models relevant to climate dynamics: {three reduced-order models of increasing complexity and a 2D Navier-Stokes model representing a turbulent flow with a localized perturbation.} In all cases, the approach accurately captures strongly nonlinear and non-Gaussian features of the system’s response, significantly outperforming traditional Gaussian approximations.
\end{abstract}

\section*{Significance Statement}
Predicting how complex stochastic systems respond to small external perturbations is central in physics, climate science, and engineering. We combine the Generalized Fluctuation–Dissipation Theorem with score-based generative modeling to accurately capture mean and higher-order (variance, skewness, kurtosis) responses. {Our framework demonstrates significant flexibility, providing accurate results for both well-behaved reduced-order models and for high-dimensional turbulent systems where such reductions are not possible.} This work provides a powerful, data-driven tool for understanding how small forcings alter the full probability distribution, including the statistics of extreme events, in strongly non-Gaussian systems.

\noindent\textbf{Classification:} Applied Mathematics.\\
\textbf{Keywords:} fluctuation–dissipation | score-based generative modeling | reduced‑order models | climate dynamics | probability response

\section{Introduction}

Understanding how a physical system responds to external perturbations is a central challenge in physics, climate science, and engineering \citep{strogatz2018nonlinear, palmer2001stochastic, baiesi2013update}. In many applications, one is not only interested in the shift of the system’s \emph{mean} state under a small forcing but also in changes in its \emph{higher-order moments}---such as variance, skewness, and kurtosis. Capturing responses in higher-order moments is crucial for reconstructing the perturbed steady-state probability distribution and for building a principled, causal framework to analyze how perturbations shape the system’s overall response, including distributional tails \citep{gfd_climate_response}. For instance, in climate science, assessing the full redistribution of probability (e.g., changes in the frequency of heat waves or cold spells) is as crucial as estimating an average temperature increase \citep{ipcc2013climate}.\\

A key theoretical tool in this context is the \emph{Generalized Fluctuation-Dissipation Theorem} (GFDT), which relates the steady-state probability density function (PDF) of a system to its linear response under a time-dependent \textit{external} perturbation \citep{giorgini_response_theory,cooper2011climate,baldovin2020understanding,ghil2020physics,majda_climate_response,MajdaBook}. In principle, GFDT offers a route to predict how the entire distribution of an observable changes without explicitly perturbing the system. However, practical applications of the GFDT are hindered by the difficulty of accurately estimating the full, often high-dimensional and non-Gaussian, steady-state PDF. In many cases the \emph{Gaussian} approximation---where the system’s PDF is replaced by a Gaussian distribution sharing the same mean and covariance---is employed. It has been shown empirically that the Gaussian approximation has high-skill in predicting responses in the mean even in nonlinear systems (e.g. \cite{GRITSUN,GERSHGORIN20101741,baldovin2020understanding}) but  introduces systematic biases for higher-order moments \citep{majda_fdt,majda_chaos}.

Recent advances in data-driven methodologies offer promising alternatives. {For instance, operator-theoretic frameworks connect a system's response to the spectral properties of its evolution operator. This viewpoint allows for expressing the response formula in terms of the poles and residues of the susceptibility function of the unperturbed system \citep{gutierrez2022some}, and it has led to practical data-driven methods for constructing response operators from time series \citep{chekroun2024kolmogorov}. More generally, response theory for any Markovian system can be formulated via a fluctuation-dissipation relation for the process generator \citep{lucarini2025interpretable}.} In particular, \emph{score-based generative modeling} has emerged as a powerful approach for sampling from complex distributions by learning the gradient of the log-PDF (the \emph{score function}) directly from data, thereby circumventing the need for full density estimation \citep{song_sde}. Moreover, modern clustering-based algorithms have shown that statistical properties of high-dimensional systems can be estimated more reliably from coarse-grained observational data than the systems' detailed dynamical trajectories \citep{falasca2024data, falasca2025FDT, giorgini2025learning, giorgini2024reduced, souza2024representing_a, souza2024representing_b, souza2024modified}.

These data-driven advances are particularly valuable in settings where directly solving the full partial differential equations (PDEs) governing a system's dynamics is computationally prohibitive or even unknown. This limitation has motivated the development of reduced-order models, which aim to reproduce the essential statistical behavior of complex systems at a fraction of the computational cost. {In many high-dimensional dynamical systems, the vast majority of the system's energy and variability is often concentrated in a small number of large-scale, coherent patterns, commonly identified through empirical orthogonal functions (EOFs) or other dimensionality reduction techniques. These dominant patterns capture the essential dynamics of the system while filtering out higher-frequency, smaller-scale fluctuations. One can identify these dominant patterns and build a stochastic model for their dynamics alone, effectively reducing the dimensionality of the problem while retaining the key physical processes that govern the system's behavior.} In climate science, in particular, stochastic reduced-order models have played a central role in both understanding and forecasting system behavior \citep{Hasselmann1976,Penland89,Penland95,MTV1,MTV2,KONDRASHOV201533,STROUNINE2010145,LucariniChekroun2023}. For example, reduced Markovian models have been successfully employed to derive effective dynamics for slow variables in multiscale systems \citep{KRAVTSOV,majda_applied_math, nan2019, keyes2023stochastic, giorgini2022non, giorgini2025data}. 

{Similar reduced-order modeling approaches have proven essential across diverse research areas dealing with nonequilibrium systems, including turbulent fluid dynamics \citep{berkooz1993proper, rowley2017model}, plasma physics \citep{peerenboom2015dimension, farcas2024scientific}, biological systems \citep{brunton2016discovering, champion2019data}, chemical reaction networks \citep{valorani2006automatic, lu2008strategies}, and neuroscience \citep{brunton2016extracting, proctor2016dynamic}.}

These approaches facilitate our understanding of high-dimensional dynamical systems by focusing on core processes while also improving computational efficiency. Furthermore, reduced-order models also shed light on how non-Gaussian features and intermittent behavior emerge from interactions between resolved and unresolved scales, \citep{falasca2025neuralmodelsmultiscalesystems}. 

{However, constructing accurate response functions for reduced-order models (ROMs) presents inherent limitations. For the GFDT to yield an accurate response from a ROM, a specific set of conditions should be met: (i) the retained slow modes must be dynamically well separated from the fast, truncated modes which will be treated as noise; and (ii) the external perturbation should project primarily onto the slow, coherent modes included in the ROM. When these conditions are violated, the ROM may fail to capture the full system's response to perturbations. Nevertheless, many problems of practical interest fall into this favorable scenario, such as studying the climate's response to large-scale forcings (e.g., uniform CO$_2$ increase) that naturally align with the system's dominant coherent modes \citep{marvel2015implications}.
}

Building on these ideas, recent studies have combined the GFDT framework with generative score-based techniques to improve predictions of the mean response in high-dimensional systems \citep{giorgini_response_theory}. In this paper, we build on the framework presented in \citep{giorgini_response_theory} by constructing \emph{higher‑order} response functions for nonlinear reduced‑order stochastic models relevant to climate science. Specifically, we examine the response of the mean (first moment) and the second, third, and fourth central moments of an observable to small external perturbations. By incorporating information from these higher-order moments, our approach provides an accurate estimate of the perturbed steady-state PDF, which is crucial for understanding changes in variability and the occurrence of extreme events. We test the approach on the following reduced-order stochastic models of increasing complexity: a one-dimensional stochastic model for low‑frequency climate variability, a slow–fast triad model designed to capture key features of the El Niño-Southern Oscillation (ENSO), and a six-dimensional stochastic barotropic model that reproduces atmospheric regime transitions \citep{charneyDeVore,de1988low, crommelin2004mechanism}. 
{We also study the case where the applied perturbation couples with all modes of the high-dimensional system. In this latter case, approximating the system's behaviour with a ROM is not possible. We show how our method can be generalized to this case by considering as an example a 2D Navier-Stokes model with a localized perturbation.} Our results demonstrate that the new method can accurately reproduce the steady-state PDFs and outperforms traditional Gaussian approximations in capturing higher-order moment responses.

The remainder of the paper is organized as follows. {In Section~\ref{sec:Methods}, we review the derivation and practical implementation of the GFDT and introduce the two complementary, data-driven techniques used for score function estimation: a clustering-based algorithm (KGMM) for low-dimensional systems, and a denoising score-matching approach for high-dimensional systems. To keep the main text focused, the methodological details are outlined in the Appendix. In Section~\ref{sec:results}, we present numerical experiments on several models of increasing complexity—the three reduced-order climate models and a high-dimensional 2D Navier-Stokes model—highlighting the improved skill of our approach in capturing higher-order responses compared to the standard Gaussian approximation.} Section~\ref{sec:limits} discusses the practical considerations and limitations of the framework, contrasting the conditions under which reduced-order models are applicable with scenarios requiring direct estimation on the full state space. Finally, Section~\ref{sec:conclusions} summarizes our findings and discusses directions for future work.

\section{Methods}
\label{sec:Methods}
In this section, we present the theoretical and computational foundations of our approach. We begin by deriving the Generalized Fluctuation-Dissipation Theorem (GFDT), which forms the basis for computing response functions in nonlinear stochastic systems. {We then present two complementary, data-driven techniques for estimating the score function, which is central to implementing GFDT: (i) the KGMM (K-means Gaussian Mixture Modeling) algorithm \citep{giorgini2025kgmm}, used for \emph{low-dimensional systems} (our scalar, triad, and barotropic ROMs); and (ii) denoising score matching, used for \emph{higher-dimensional systems} (e.g., the 2D Navier--Stokes vorticity field on a $32\times 32$ grid). Throughout this work we follow the rule of thumb: lower-dimensional in state space $\rightarrow$ KGMM with a multilayer perceptron architecture; higher-dimensional in state space $\rightarrow$ denoising score-matching with a U\,-Net architecture. See Appendix~\S\,6.3 for detailed guidance and a practical comparison.} Additional details on the maximum entropy framework used to reconstruct perturbed steady-state distributions from estimated moment responses are provided in the Appendix ({Appendix~\S\,1}).

\subsection{The Generalized Fluctuation-Dissipation Theorem}
\label{subsec:GFDT}

We consider a stochastic dynamical system governed by
\begin{equation}
\dot{\bm{x}}(t) = \bm{F}(\bm{x}) + \bm{\sigma}\,\bm{\xi}(t),
\end{equation}
where $\bm{x} \in \mathbb{R}^n$, $\bm{F}(\bm{x})$ is a deterministic drift term, $\bm{\sigma}$ is the noise amplitude matrix, and $\bm{\xi}$ is a standard Gaussian white noise process. {The noise amplitude matrix $\bm{\sigma}$ can be constant, corresponding to additive noise, or a state-dependent function $\bm{\sigma}(\bm{x})$, corresponding to multiplicative noise. In the latter case, the equation is interpreted in the It\^{o} sense, a choice reflected in the form of the Fokker-Planck operator below. Our framework relies on the assumption that the system is ergodic, admitting a unique and smooth invariant measure $\rho_S(\bm{x})$. Sufficient conditions for this are, for instance, the hypoellipticity of the Fokker-Planck operator and the existence of a Lyapunov function that ensures the confinement of trajectories \citep{pavliotis2014stochastic}.

The system evolves towards a steady state described by a probability density $\rho_S(\bm{x})$, which satisfies the stationary Fokker–Planck equation:
\begin{equation}
\mathcal{L}_0\,\rho_S(\bm{x}) = 0,
\end{equation}
where the Fokker–Planck operator $\mathcal{L}_0$ is defined as
\begin{equation}
\mathcal{L}_0\,\rho(\bm{x}) = -\nabla\cdot\big(\bm{F}(\bm{x})\,\rho(\bm{x})\big) + \frac{1}{2}\nabla\nabla^\top : \big(\bm{D}\,\rho(\bm{x})\big),
\end{equation}
with $\bm{D} = \bm{\sigma}\bm{\sigma}^\top$ the diffusion matrix, and $:\,$ denoting the double contraction of tensors.

We now introduce a small, time-dependent perturbation that factorizes into a small spatial component $\bm{u}(\bm{x})$ and a temporal modulation $f(t)$, so that the dynamics become
\begin{equation}
\dot{\bm{x}}(t) = \bm{F}(\bm{x}) + \bm{u}(\bm{x})\,f(t) + \bm{\sigma}\,\bm{\xi}(t).
\end{equation}
This perturbation induces a small variation in the probability density, denoted $\delta\rho(\bm{x},t)$, which satisfies, to first order in $\bm{u}(\bm{x})$,
\begin{equation}
\frac{\partial \delta\rho(\bm{x},t)}{\partial t} = \mathcal{L}_0\,\delta\rho(\bm{x},t) + f(t)\,\mathcal{L}_1\,\rho_S(\bm{x}),
\label{eq:delta_rho}
\end{equation}
with the perturbation operator $\mathcal{L}_1$ defined by
\begin{equation}
\mathcal{L}_1\,\rho(\bm{x}) = -\nabla\cdot\big(\bm{u}(\bm{x})\,\rho(\bm{x})\big).
\end{equation}
For the perturbed density $\rho_S + \delta\rho$ to remain a valid probability distribution, the total probability must be conserved. This requires the integral of the density variation over the entire state space to be zero at all times. This consistency constraint, $\int \delta\rho(\mathbf{x},t)\,d\mathbf{x} = 0$, is satisfied by Equation \eqref{eq:delta_rho}, given that the Fokker-Planck operator $\mathcal{L}_0$ preserves normalization (i.e., its integral over the domain is zero, assuming no-flux boundary conditions) and the system is unperturbed at $t=0$ (see {Appendix~\S\,2}.)

By formally integrating this linearized equation (see {Appendix~\S\,3}), the resulting first-order correction to the expectation value of any observable $A(\bm{x})$ is given by
\begin{equation}
\langle \delta A(t) \rangle = \int_0^t \!\! \, f(t')\,\big\langle A(\bm{x}(t))\,B(\bm{x}(t'))\big\rangle_0 \mathrm{d}t' = \int_0^t \bm{R}(t - t')\,f(t')\,\mathrm{d}t',
\label{eq:response}
\end{equation}
where the conjugate observable $B(\bm{x})$ is
\begin{equation}
B(\bm{x}) = \frac{\mathcal{L}_1\,\rho_S(\bm{x})}{\rho_S(\bm{x})} = -\frac{\nabla\cdot\big(\bm{u}(\bm{x})\,\rho_S(\bm{x})\big)}{\rho_S(\bm{x})},
\end{equation}
and $\bm{R}(t) = \langle A(\bm{x}(t))\,B(\bm{x}(0))\rangle_0$ is the linear response function, by definition equal to the response to an impulse (delta) perturbation $f(t) = \delta(t)$ \cite{Risken}. $\langle \cdot \rangle_0$ denotes the expectation with respect to the unperturbed steady-state distribution $\rho_S(\bm{x})$.

Applying the product rule, we obtain
\begin{equation}
\nabla\cdot\big(\bm{u}(\bm{x})\,\rho_S(\bm{x})\big) = \rho_S(\bm{x})\,\nabla\cdot \bm{u}(\bm{x}) + \bm{u}(\bm{x})\cdot\nabla\rho_S(\bm{x}),
\end{equation}
and thus the conjugate observable takes the explicit form
\begin{equation}
B(\bm{x}) = -\nabla\cdot \bm{u}(\bm{x}) - \bm{u}(\bm{x})\cdot \nabla \ln\rho_S(\bm{x}).
\label{eq:B}
\end{equation}

{The structure of the conjugate observable $B(\bm{x})$ depends critically on the nature of the perturbation $\bm{u}(\bm{x})$. The expression simplifies considerably for a state-independent forcing, where $\bm{u}(\bm{x})$ is a constant vector. In this scenario, often employed in idealized models \citep{majda2009} or certain diagnostic frameworks \citep{Penland95}, the divergence term vanishes ($\nabla\cdot \bm{u}(\bm{x})=0$), and $B(\bm{x})$ becomes proportional to the score function, $-\nabla \ln\rho_S(\bm{x})$. In contrast, physically realistic forcings in climate science cab be inherently state-dependent. For example, the climatic forcing associated with an increase in CO$_2$ concentration depends non-linearly on the state of the atmosphere (e.g., its temperature and water vapor content), meaning $\bm{u} = \bm{u}(\bm{x})$. For such cases, the divergence term $\nabla\cdot\bm{u}(\bm{x})$ is non-zero and physically crucial, making the full expression for $B(\bm{x})$ essential for an accurate response calculation \citep{lucarini2017predicting, lembo2020beyond}.}

Often the unperturbed steady state is approximated with a multivariate Gaussian. In this case we have
\begin{equation}
\rho_S(\bm{x}) = \frac{1}{\sqrt{(2\pi)^n \det\bm{\Sigma}}}\,\exp\!\left[-\frac{1}{2}(\bm{x} - \bm{\mu})^\top \bm{\Sigma}^{-1}(\bm{x} - \bm{\mu})\right],
\end{equation}
{where $\bm{\mu}$ and $\bm{\Sigma}$ respectively represent the mean and covariance matrix of the system. In this case we have:}
\begin{equation}
\ln \rho_S(\bm{x}) = -\frac{1}{2}(\bm{x} - \bm{\mu})^\top \bm{\Sigma}^{-1}(\bm{x} - \bm{\mu}) + \text{const},
\end{equation}
and consequently,
\begin{equation}
\nabla \ln \rho_S(\bm{x}) = -\bm{\Sigma}^{-1}(\bm{x} - \bm{\mu}).
\end{equation}
Substituting into the expression for $B(\bm{x})$, we obtain
\begin{equation}
B(\bm{x}) = -\nabla\cdot \bm{u}(\bm{x}) + \bm{u}(\bm{x})^\top \bm{\Sigma}^{-1}(\bm{x} - \bm{\mu}).
\end{equation}
In this case, it is straightforward to verify that, under external forcings $\bm{u}(\bm{x})_i = \bm{e}_i$ the response function reduces to $\bm{R}(t) = \langle A(\bm{x}(t)) \bm{\Sigma}^{-1}(\bm{x} - \bm{\mu})(0) \rangle_0$, as commonly adopted in climate studies \cite{MajdaBook}. Note that while the analytical score is approximated using that of a Gaussian distribution, the ensemble averages $\langle \cdot \rangle_0$ are still obtained using the original data and therefore over the system's steady state distribution. This approach differs from linear inverse model strategies (e.g., \cite{Penland89}), which assume a time-invariant Gaussian measure from the outset. Such models would yield zero response to any quadratic functional, such as the variance (see \cite{majda2009,falasca2025neuralmodelsmultiscalesystems}). Because of this distinction, the approximation discussed here is sometimes referred to as a quasi-Gaussian approximation in the literature. For simplicity, we will simply refer to it as the Gaussian approximation throughout this work.

{
\subsection{Score Function Estimation via KGMM}
\label{subsec:KGMM}

To apply the Generalized Fluctuation-Dissipation Theorem (GFDT), we need the score function, $\nabla \ln \rho_S(\bm{x})$, which is the gradient of the logarithm of the system's steady-state distribution. Since this function is rarely known analytically, it must be estimated from data. For systems where the dynamics can be effectively described in a low-dimensional space {(as in our scalar, triad, and barotropic ROMs)}, we employ the KGMM (K-means Gaussian Mixture Modeling) algorithm, a hybrid statistical-learning method designed for efficient and accurate score estimation \citep{giorgini2025kgmm}.

The KGMM method computes the score by leveraging a probabilistic identity that avoids the numerical instabilities associated with direct differentiation of a data-based density estimate. The procedure begins by creating a new set of ``perturbed'' samples, where each original data point is shifted by a small, random vector drawn from a Gaussian distribution. {This construction is closely connected to denoising score matching: training a network on perturbed samples to predict the added noise is equivalent to learning the score of a Gaussian–smoothed density (i.e., a Gaussian mixture with isotropic kernels); see \citep{kamb2024analytic, baptista2025memorization}.} The algorithm's core insight is that the score function is directly proportional to the conditional expectation—or local average—of these displacements. In practice, this is achieved by partitioning the perturbed data samples into clusters and computing the average displacement within each cluster, which provides a discrete estimate of the score at the cluster centroids. Finally, a neural network is trained to interpolate these discrete estimates, yielding a continuous representation of the score function over the entire domain.

{Crucially, both KGMM and denoising score-matching is introduced to avoid the numerical instabilities of the direct GMM score estimator—differentiating a Gaussian-mixture density becomes ill-conditioned and noise-amplifying as the kernel width $\sigma_G \to 0$—by recasting score evaluation as a stable conditional-expectation regression (see Appendix~\S\,6.1).}

A detailed description of the algorithm, including the governing equations, implementation steps, and hyperparameter choices, is provided in {Appendix~\S\,6.1}.

\subsection{Score Function Estimation via Denoising Score-Matching}
\label{subsec:UNet}

When a system's dynamics cannot be effectively reduced to a low-dimensional model—for instance, if there is no clear separation of timescales or if a perturbation couples with all modes of a high-dimensional system—the KGMM clustering algorithm becomes computationally prohibitive. In these scenarios, we estimate the score function directly on the full, high-dimensional state space.

For this purpose, we employ a denoising score matching approach, training a neural network to learn the score directly from data \citep{useful_diffusion}. This method circumvents explicit clustering by framing the problem as one of learning to denoise perturbed samples. We use a U-Net architecture, which is particularly well-suited for spatially-extended systems due to its encoder-decoder structure with skip connections that preserve spatial information across multiple scales. The network is trained to minimize a denoising score matching loss function. We use the same methodology as what was implemented in 
\citep{giorgini_response_theory} with appropriate modifications to higher-order responses.  This technique provides a scalable and powerful alternative for score estimation in high-dimensional settings where reduced-order modeling is not applicable. The full details of the U-Net architecture, loss function, and training procedure are available in {Appendix~\S\,6.2}.
}
\section{Results}
\label{sec:results}

{In what follows, we estimate the mean (first moment) and the second, third, and fourth central moments of the perturbed steady‑state distribution for four representative nonlinear stochastic systems of increasing complexity: three reduced-order models (a scalar stochastic model for low‑frequency climate variability, a slow‑fast triad model for ENSO, and a six‑dimensional barotropic model for atmospheric regimes), and one high-dimensional partial differential equation (a 2D Navier-Stokes model for turbulent flow).} These systems serve as ideal testbeds for evaluating the accuracy of different approximations to the Generalized Fluctuation-Dissipation Theorem (GFDT) in predicting higher-order responses under small external perturbations.

{For each system, we construct the response functions as defined in ~\eqref{eq:response}. Specifically, we focus on response functions for the mean (first moment) and the second, third, and fourth central moments of the invariant distribution by choosing as observables \(\bm{x}\), \((\bm{x}-\bm{\mu})^2\), \((\bm{x}-\bm{\mu})^3\), and \((\bm{x}-\bm{\mu})^4\), where \(\bm{x}\) denotes the system state and \(\bm{\mu}\) its unperturbed mean. {Additionally, for the 2D Navier–Stokes example, we also analyze the fifth central moment \((\bm{x}-\bm{\mu})^5\).} We assess three distinct approaches: (i) a data-driven method that constructs the response using the score function estimated via either the KGMM algorithm for the low-dimensional models or a U-Net for the high-dimensional system; (ii) a Gaussian approximation, which assumes the steady-state distribution to be multivariate Gaussian, fully characterized by its mean and covariance; and (iii) a reference or ``ground truth'' response. In the case of the scalar model, the ground truth has been derived analytically in \cite{NormalForms} from the known stationary distribution. For the other models, where an analytic form of the score is not available, the ground truth response is computed numerically by performing ensemble integrations of both the unperturbed and perturbed systems and evaluating the change in the moments, see for example the \cite{giorgini_response_theory,falasca2025neuralmodelsmultiscalesystems} as well as the Appendix of \cite{boffett}. In all examples, we normalize the data such that the unperturbed time series in each dimension has zero mean and unit variance. This preprocessing step facilitates the comparison across variables and models, and ensures that the Gaussian approximation always corresponds to a standard normal distribution. For systems subject to a state-independent perturbation, as in the first and third examples, we normalize the response functions by the perturbation amplitude, rendering them independent of its magnitude. }

In settings where direct computation of response functions is infeasible—either because the underlying equations are unknown or integration is prohibitively expensive—the only reliable way to validate the {estimated score} (and, by extension, any response predictions) is to compare observed and {generative model-based} marginal PDFs. Concretely, we simulate the Langevin dynamics
\begin{equation}
\dot{\bm{x}}(t) \;=\; \bm{s}(\bm{x}(t)) \;+\; \sqrt{2}\,\bm{\xi}(t),
\label{eq:langevin}
\end{equation}
where \(\bm{s}(\bm{x}) = \nabla \ln \rho_S(\bm{x})\) is the score learned via {our data-driven methods}. The degree to which the simulated marginals reproduce the empirically observed steady‐state PDFs thus provides the sole stringent diagnostic that the score has been learned accurately and that any inferred response functions can be trusted. For this reason, we have also compared the marginal steady‐state PDFs obtained from these Langevin simulations directly against the empirical distributions, offering an independent check on the fidelity of the estimated score function.

This comparative analysis allows us to quantify the extent to which generative score modeling improves the accuracy of higher-order response predictions relative to traditional linear approximations, especially in regimes characterized by strong non-Gaussianity and nonlinear interactions.

\subsection{Scalar Stochastic Model for Low-Frequency Climate Variability}
\label{sec:triad_results}

We consider a one-dimensional stochastic model for low‑frequency climate variability. The model was originally derived by  \cite{NormalForms}  using stochastic reduction techniques developed in \cite{MTV1,MTV2}. For further details on the derivation and relevant literature, we refer the reader to Section 4b of \cite{majda2009}. This reduced-order stochastic model has been successfully used to fit nonlinear scalar dynamics from low-frequency data of a general circulation model and has also served as a testbed for fluctuation-dissipation theorem (FDT) analyses, see for example \citep{majda2009,majdaStructuralStability} in the context of the Gaussian approximation. The model is given by the scalar nonlinear stochastic differential equation:
\begin{equation}
\dot{x}(t) = F + a x(t) + b x^2(t) - c x^3(t) + \sigma_1\,\xi_1(t) + \sigma_2(x)\xi_2(t),
\end{equation}
where $\xi_1(t)$ and $\xi_2(t)$ are independent standard white noise processes. In the present work, we consider a closely related scalar model introduced in \citep{nan2019}, which builds upon similar principles but features no multiplicative noise, that is, $\sigma_2(x)=0$. This leads to a simplified, purely additive stochastic model of the form:
\begin{equation}
\dot{x}(t) = F + a x(t) + b x^2(t) - c x^3(t) + \sigma\,\xi(t),
\end{equation}
where $\xi(t)$ is standard white noise. {These dynamics can be interpreted as the motion of an overdamped particle in a one-dimensional potential energy landscape $U(x)$, where the deterministic force is given by $F(x) = -dU/dx$. The potential is a quartic polynomial of the form $U(x) = - (Fx + \frac{a}{2}x^2 + \frac{b}{3}x^3 - \frac{c}{4}x^4)$.} \\


The coefficients of the model used in this work are listed in Table~1 of {Appendix~\S\,5}. The input parameters of the KGMM algorithm (see Section~\ref{subsec:KGMM}) are set to $\sigma_G = 0.05$ and $N_C = 348$.

A key motivation for considering this model is the availability of an analytical expression for the score function, defined as the gradient of the logarithm of the stationary distribution:
\begin{equation}
s(x) = \frac{d}{dx} \log \rho_S(x) = \frac{2}{\sigma^2} \left( F + a x + b x^2 - c x^3 \right).
\label{eq:score_function}
\end{equation}

Correspondingly, the stationary probability density function (PDF) of the system can be written in closed form as:
\begin{equation}
\rho_S(x) = \mathcal{N}^{-1} \exp\left[ \frac{2}{\sigma^2} \left( Fx + \frac{a}{2} x^2 + \frac{b}{3} x^3 - \frac{c}{4} x^4 \right) \right],
\label{eq:pdf}
\end{equation}
where $\mathcal{N}$ is a normalization constant ensuring that $\rho_S(x)$ integrates to 1.\\

We now construct impulse response functions through GFDT as $R(t) = \langle A(x(t))\,B(x(0))\rangle_0$, and remind the reader that $\langle \cdot \rangle_0$ denotes the expectation with respect to the unperturbed steady-state distribution $\rho_S(x)$. We consider the simple case where the conjugate observable is defined as $B(x) = - s(x) = - \frac{d}{dx} \log \rho_S(x)$. To estimate the score function $s(x)$ we consider three different strategies: (i) analytically from ~\eqref{eq:score_function}, (ii) using the KGMM method, and (iii) under the Gaussian approximation. In Figure~\ref{fig:reduced_responses}, we compare the unperturbed steady-state PDF as well as the first four moment response functions obtained with these three approaches. The results show that the score function estimated via KGMM yields a nearly perfect match with the analytic response, both in terms of the steady‑state PDF and the responses of the mean (first moment) and the second, third, and fourth central moments. This confirms the ability of KGMM to accurately reconstruct the score function. {As expected from previous studies (e.g. \citep{baldovin2020understanding}), the Gaussian approximation provides a very good estimation for changes in the mean. For the higher-order moments, it can sometimes capture the qualitative nature of the response, such as the initial sign and relaxation timescale for the variance. However, it introduces significant quantitative biases and becomes increasingly unreliable for higher orders, reflecting its inability to capture the non-Gaussian features of the underlying dynamics.}

\begin{figure*}
    \centering
    \includegraphics[width=\textwidth]{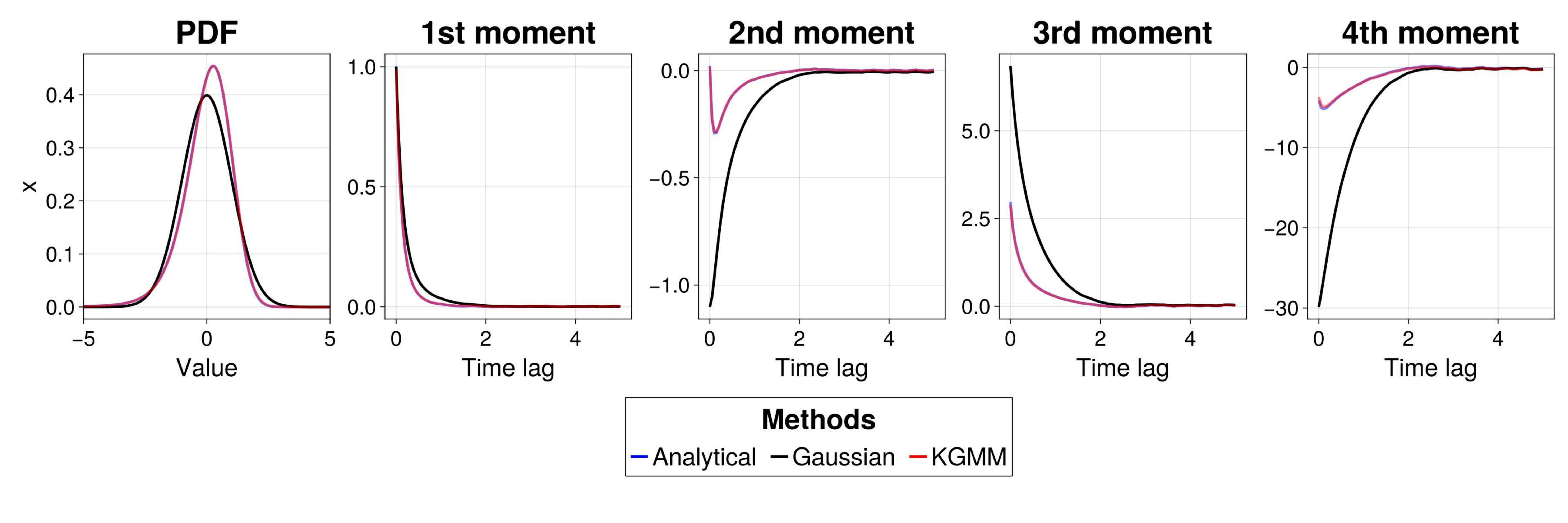}
    \caption{Left: True unperturbed PDF (blue) compared with a normal Gaussian (black) and the PDF obtained integrating ~\eqref{eq:langevin} with the KGMM score function (red). Right: Response functions predicted via GFDT. The true response functions (blue) are obtained using the analytic score function and are compared with the ones obtained using the KGMM score function (red) and the linear approximation (black). {Note that the non-zero response of the second moment under the Gaussian approximation is due to the quasi-Gaussian framework used (see Section 2.1), where the expectation is taken over the true non-Gaussian distribution.}}
    \label{fig:reduced_responses}
\end{figure*}

\subsection{Slow-Fast Triad Model and Application to ENSO}
\label{sec:enso_results}

The second test case considers the slow-fast triad model from \citep{thual2016simple}:
\begin{equation}
\begin{aligned}
\dot{u_1} &= -d_{u}u_1 - \omega u_2 + \tau + \sigma_{u1}\,\xi_1(t),\\
\dot{u_2} &= -d_{u}u_2 + \omega u_1 + \sigma_{u2}\,\xi_2(t),\\
\dot{\tau} &= -d_{\tau}\,\tau + \sigma_{\tau}(u_1)\,\xi_3(t),
\end{aligned}
\end{equation}
with parameters listed in Table~2 of {Appendix~\S\,5}. The input parameters of the KGMM algorithm (see Section~\ref{subsec:KGMM}) are set to $\sigma_G = 0.05$ and $N_C = 7353$.

This model is designed to mimic multi-scale interactions characteristic of the El Niño-Southern Oscillation (ENSO), where the slow variables $u_1$ and $u_2$ represent the large-scale ocean-atmosphere state, and the fast variable $\tau$ represents rapid wind bursts. Differently from the previous Section, here we are interested in studying the system's response to perturbations in the damping coefficient of the fast variable $\tau$. Specifically, the perturbation is applied as:
\begin{equation}
\delta d_{\tau} = -0.2 d_{\tau} = -0.4.
\end{equation}
This modification increases the memory time of $\tau$ (the wind bursts), thereby enhancing its strength. Physically, this perturbation is interpreted as an increase in Madden-Julian oscillation or monsoon activity in the Western Pacific, which in turn modifies the behavior of ENSO. A direct consequence of this perturbation is an increase in the variance of $u_1$ and $u_2$, leading to a higher occurrence of strong ENSO events.\\

In this case, the conjugate observable needs to be considered in its full formulation as $B(\bm{x}) = -\nabla\cdot \bm{u}(\bm{x}) - \bm{u}(\bm{x})\cdot \nabla \ln\rho_S(\bm{x})$. Unlike the triad model, no analytic expression for the steady-state distribution or score function is available for this system. Therefore, the ground truth response was computed by directly simulating an ensemble of both unperturbed and perturbed systems and evaluating the change in moments (see e.g. \cite{giorgini_response_theory}). These serve as a benchmark for validating the GFDT-based predictions. As in the previous example, we compared the response of the mean (first moment) and the second, third, and fourth central moments of each variable using three methods: GFDT with the KGMM-estimated score function, GFDT under the Gaussian linear approximation, and the numerical ground truth. We also compared the unperturbed marginal steady-state PDFs predicted by KGMM to the empirically observed distributions by integrating the Langevin equation in \eqref{eq:langevin}.\\

The results are reported in Figure~\ref{fig:enso_responses}. In all three dimensions, the score function estimated by KGMM allows the GFDT framework to accurately reproduce the response functions and the steady-state PDFs. The agreement with the numerically computed ground truth is excellent, indicating that the KGMM method successfully reconstructs the non-Gaussian statistical structure of the system. In contrast, the linear approximation yields visible discrepancies in all moments, especially for the strongly skewed and heavy-tailed distribution of $\tau$.

\begin{figure*}
    \centering
    \includegraphics[width=\textwidth]{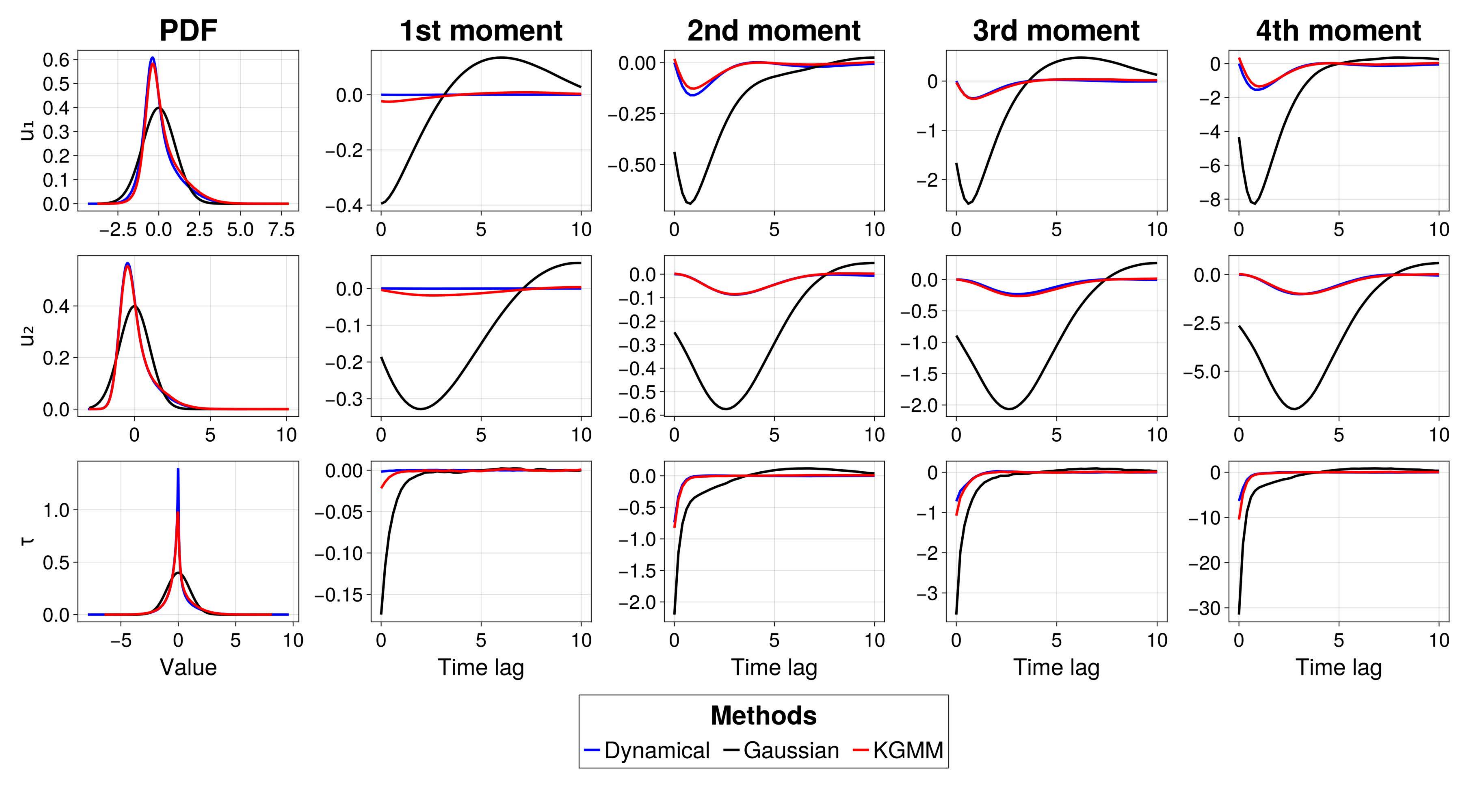}
    \caption{Comparison of unperturbed marginal steady-state PDFs (leftmost column) and responses of the first four moments (remaining columns) in the slow-fast triad model for variables $u_1$, $u_2$, and $\tau$ (rows). Response functions are shown as a function of time lag. Predictions obtained via GFDT using the KGMM-estimated score function (red), the Gaussian approximation (black), and numerical ground truth (blue) are reported.}
    \label{fig:enso_responses}
\end{figure*}

\subsection{Stochastic Barotropic Model for Atmospheric Regime Transitions}

The large-scale circulation of the atmosphere often exhibits persistent patterns, such as zonal and blocked flow regimes, with abrupt transitions between them. These features can be captured through low-order models that isolate the essential mechanisms responsible for regime persistence and transitions \citep{charneyDeVore}. The six-dimensional model considered here is based on a Galerkin truncation of the barotropic vorticity equation on a $\beta$-plane channel with topography. The original formulation was introduced by Charney and DeVore \citep{charneyDeVore}, and the specific version used here follows a slightly modified form proposed by De Swart \citep{de1988low}. This model has been widely studied, particularly in the context of stochastic regime dynamics \citep[e.g.,][]{crommelin2004mechanism, dorrington2023interaction, grafke2019numerical}. It retains two zonal modes and two pairs of wave modes, enabling nonlinear interactions between the waves and the mean flow, as well as capturing the influence of topographic forcing. Physically, the model describes the evolution of the barotropic streamfunction field, subject to planetary rotation, Newtonian damping toward a prescribed zonal background state, and orographic forcing. The six prognostic variables represent amplitudes of selected Fourier modes, and the equations contain linear terms corresponding to damping and rotation, as well as quadratic nonlinearities encoding advective interactions. The system supports multiple equilibria and intermittent transitions between weakly chaotic and more persistent quasi-stationary states. By adding stochastic forcing, one can probe how unresolved processes affect the stability and persistence of regimes \citep{dorrington2023interaction}. The stochastic version of the model takes the form:

\begin{equation}
\begin{aligned}
\dot{x}_1 &= \tilde{\gamma}_1 x_3 - C(x_1 - x_1^*) + \sigma \xi_1(t), \\
\dot{x}_2 &= -(\alpha_1 x_1 - \beta_1)x_3 - C x_2 - \delta_1 x_4 x_6 + \sigma \xi_2(t), \\
\dot{x}_3 &= (\alpha_1 x_1 - \beta_1)x_2 - \gamma_1 x_1 - C x_3 + \delta_1 x_4 x_5 + \sigma \xi_3(t), \\
\dot{x}_4 &= \tilde{\gamma}_2 x_6 - C(x_4 - x_4^*) + \varepsilon (x_2 x_6 - x_3 x_5) + \sigma \xi_4(t), \\
\dot{x}_5 &= -(\alpha_2 x_1 - \beta_2)x_6 - C x_5 - \delta_2 x_4 x_3 + \sigma \xi_5(t), \\
\dot{x}_6 &= (\alpha_2 x_1 - \beta_2)x_5 - \gamma_2 x_4 - C x_6 + \delta_2 x_4 x_2 + \sigma \xi_6(t).
\end{aligned}
\end{equation}

Here, $x_1$ and $x_4$ denote the amplitudes of zonal modes, while the remaining variables represent wave modes. The parameter $C$ denotes the damping rate, and $(x_1^*, x_4^*)$ are the prescribed zonal background states. Stochastic forcing enters additively through independent Gaussian white noise processes $\xi_i(t)$ with common amplitude $\sigma$.\\

The values of the coefficients used in this study are summarized in Table~3 of {Appendix~\S\,5}, corresponding to a system with channel aspect ratio $b = 1.6$, $\beta = 1.25$, and topographic forcing strength $\gamma = 0.2$. The forcing terms $x_1^* = 0.95$ and $x_4^* = -0.76095$ define the equilibrium zonal profile. The input parameters of the KGMM algorithm (see Section~\ref{subsec:KGMM}) are set to $\sigma_G = 0.05$ and $N_C = 30309$.

We now construct impulse response functions through GFDT as $\bm{R}(t) = \langle A(\bm{x}(t))\,B(\bm{x}(0))\rangle_0$, and remind the reader that $\langle \cdot \rangle_0$ denotes the expectation with respect to the unperturbed steady-state distribution $\rho_S(\bm{x})$. As in Section \ref{sec:triad_results} we consider the case where $B(\bm{x}) = - \bm{s}(\bm{x}) = - \nabla \log \rho_S(\bm{x})$. To estimate the score function $\bm{s}(\bm{x})$ we consider three different strategies: the Gaussian linear approximation, the KGMM-estimated score function, and direct numerical simulations of the perturbed system, which serve as the ground truth. Figure~\ref{fig:CdV_responses} shows the results for all six state variables of the model, comparing the unperturbed steady-state PDFs (leftmost column) and the temporal evolution of the response of the mean (first moment) and the second, third, and fourth central moments (remaining columns) for a state-independent, impulse perturbation applied to $x_1$.

The KGMM-based GFDT predictions exhibit remarkable agreement with the numerical results across all variables and moment orders. The Gaussian approximation shows good estimates for changes in the mean but significant deviations for higher order moments. This is especially true for responses in the perturbed variable $x_1$, where the effects of the perturbation are most pronounced. These discrepancies confirm the limitations of the Gaussian approximation in representing nonlinear and asymmetric responses, which are typical in regime-transition dynamics such as blocking or wave-mean flow interactions. Notably, the KGMM approach captures both sharp peaks and heavy tails in the marginal distributions, which are characteristic of systems with intermittent transitions.

\begin{figure*}
    \centering
    \includegraphics[width=\textwidth]{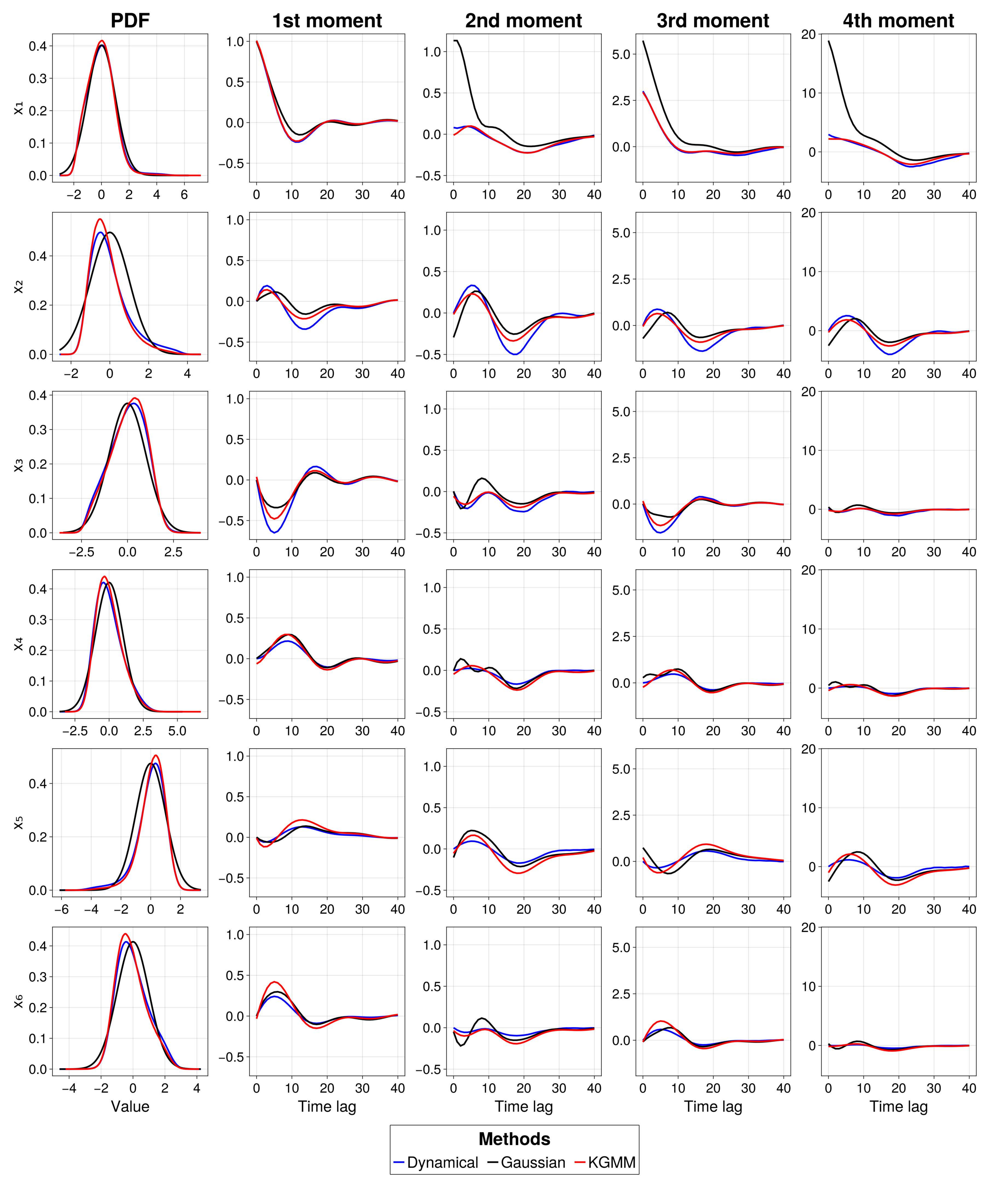}
    \caption{Comparison of the unperturbed marginal steady-state PDFs (left column) and response functions of the first four moments (remaining columns) for all six variables in the stochastic barotropic model. Results are shown for GFDT with score function estimated via KGMM (red), the linear Gaussian approximation (black), and numerical ensemble simulations (blue). }
    \label{fig:CdV_responses}
\end{figure*}

{
\subsection{Two-dimensional Navier-Stokes equation}
\label{subsec:NavierStokes}

As a representative example of a high-dimensional system where reduced-order modeling is not applicable, we consider the two-dimensional Navier-Stokes model as described in \citep{Flierl_Souza_2024, giorgini_response_theory}, where we take the state variable to be the vorticity $\zeta$. This system was chosen as a stepping-stone towards the more complex fluids observed in climate systems. The data-generating equation is:
\begin{align}\label{eq:navierstokes_eom}
\partial_t \zeta  &= \partial_x \psi \partial_y \zeta - \partial_x \zeta \partial_y \psi - A \partial_x \zeta +  \mathcal{D} \zeta  + \varsigma \\
\psi &= \Delta^{-1} \zeta \\
\mathcal{D}\zeta &= - \nu_h \Delta^{-2}\zeta -  \nu \Delta^2 \zeta - \nu_0 \int \zeta \,\mathrm{d} x \,\mathrm{d} y 
\end{align}
where $\mathcal{D}$ is a dissipation operator with constants $\nu_h = 10^{-2}$, $\nu = 10^{-5}$, $\nu_0 = (2\pi)^{-2}$, $A = \pi$ is a mean advection term, and $\varsigma$ is a random wave forcing. The random wave forcing introduces hidden dynamics into the system, thus the score function is estimated using only partial state information. The inverse Laplacian is defined to preserve the mean of the original variable.

The system is discretized on a $32 \times 32$ periodic grid using a pseudospectral method. Thus the system consists of $1024$ degrees of freedom in the Navier-Stokes equations. The forcing term maintains energy injection at large scales while viscous dissipation removes energy at small scales, establishing a turbulent steady state. The stochastic forcing $\varsigma$ represents small-scale fluctuations and ensures ergodicity of the system.

We consider a localized perturbation applied to the vorticity field at a single pixel in the computational domain, representing the most localized forcing possible in the discretized system. This single-pixel perturbation couples with all spatial modes of the system, preventing the use of reduced-order models for response prediction and representing an extreme case where external forcing is applied at a single spatial location.

For this high-dimensional system, we collect $N = 10^5$ samples from the unperturbed steady state and train the U-Net score estimator described above. The network is trained for 500 epochs with a batch size of 64 using the Adam optimizer. The observables we considered were the moments of the probability density function. Due to the reflection symmetry of the vorticity field in this system (where positive and negative vorticity are equally probable), the even moments exhibit zero response to the localized perturbation, as the perturbation preserves this antisymmetric property of the underlying dynamics.

First, we note that the Gaussian approximation provides a very good, first-order estimation of the linear response in the 1st and 3rd moment. This further underscores the utility of the Gaussian approximation as a baseline for more complex methodologies. Second, our results clearly demonstrate the benefit of generative modeling to estimate linear responses in higher-order moments through the GFDT and the methodology in \cite{giorgini_response_theory}. Figure~\ref{fig:navier_stokes_response} demonstrates that the U-Net approach provides significantly better agreement with the numerical ground truth compared to the linear approximation, particularly for the higher-order moments where non-Gaussian effects become more pronounced. This example demonstrates the capability of the U-Net-based approach to handle complex, high-dimensional systems where traditional reduced-order modeling fails due to the spatially localized nature of the perturbation and the absence of clear scale separation in the turbulent flow.

\begin{figure*}
    \centering
    \includegraphics[width=\textwidth]{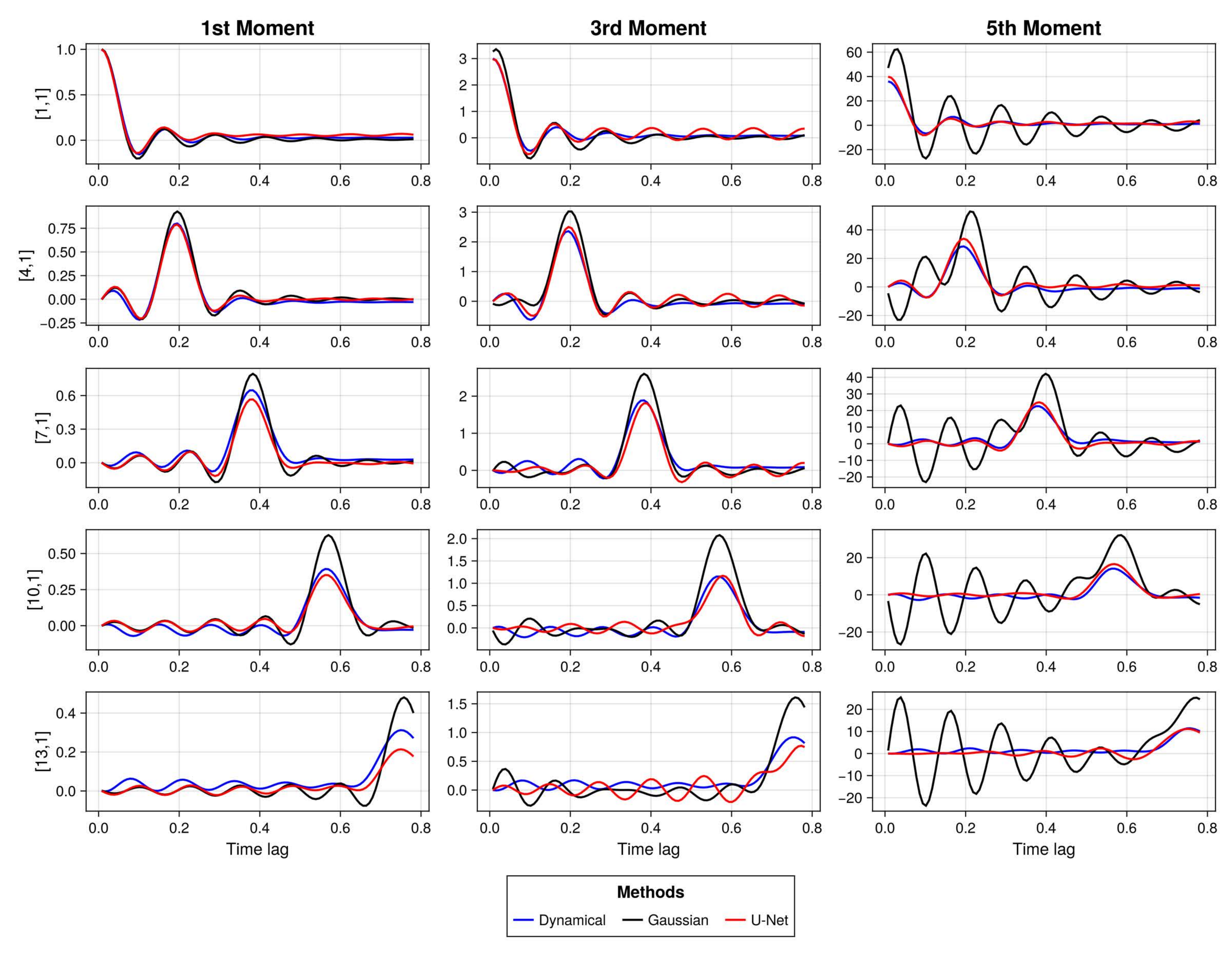}
    \caption{{Comparison of response functions for the 2D Navier-Stokes system with single-pixel perturbation. The plots show the response of the 1st, 3rd, and 5th moments across different time lags for five different observables, calculated using numerical integration (blue), U-Net score estimation (red), and linear Gaussian approximation (black). The U-Net method demonstrates superior agreement with the numerical ground truth compared to the linear approximation, particularly for higher-order moments where non-Gaussian effects are more significant.}}
    \label{fig:navier_stokes_response}
\end{figure*}
}

\section{Discussion: Limitations and practical considerations} \label{sec:limits}

Our results show that when (i) a large amount of data is available and (ii) the full state vector $\bm{x}(t)$ is known, the system’s probability density response to small external perturbations can indeed be recovered from the Generalized Fluctuation-Dissipation Theorem (GFDT), as expected from theoretical arguments \citep{nonEqStatMech}. {Specifically, we have demonstrated two complementary data-driven strategies whose selection hinges on the \emph{intrinsic spatial organization} of the data: for systems that are effectively lower-dimensional in state space (our scalar, triad, and barotropic ROMs), we estimate the score with KGMM and a multilayer perceptron; when the spatial organization is higher-dimensional (e.g., the 2D Navier--Stokes vorticity field on a $32\times32$ grid), we estimate the score with denosing score-matching and a U-Net. See {Appendix~\S\,6.3} for guidance and a side-by-side comparison, and {Appendix~\S\,6.1} and {Appendix~\S\,6.2} for method details.} The experiments presented here represent a valuable step toward an equation-free framework for studying how the probability distribution of real-world dynamical systems responds to small external perturbations.\\

However, several limitations and caveats must be considered when applying these tools to high-dimensional, ``real-world'' complex systems. We discuss such caveats below:
\begin{itemize}
    \item \textit{Small data regime.} A straightforward practical limitation arises when only relatively short time series are available. As with many machine learning-based approaches, the performance of our method depends heavily on the sample size. {In Appendix~\S\,7, we show that KGMM remains comparatively robust in lower-dimensional systems even when data are limited. In higher-than-one-dimensional settings, U-Net accuracy improves markedly with more samples; when data are scarce but PCA/EOFs capture the relevant variance, a PCA$\to$KGMM pipeline can be preferable.}
    \item \textit{Partially observed state vector.} Even with access to infinitely long time series, a more fundamental issue remains: in most real-world scenarios, we rarely observe (or even know) the full state vector of the system. Instead, we typically have access to only a small subset of variables—a projection of the full dynamics. This longstanding problem, emphasized at least since Onsager and Machlup (1953) \citep{Onsager}, remains a central concern in modern studies \citep{Cecconi,Hosni,BaldovinEntropy,baldovin2020understanding}. One might be tempted to address this challenge using the Takens embedding theorem \citep{Takens}, reconstructing the underlying dynamics from partial observations. However, this strategy is severely limited in practice: it is not applicable to stochastic systems and becomes rapidly infeasible as the system dimensionality increases; see \cite{Cecconi,BaldovinEntropy,Lucente} for an in-depth discussion. For these reasons, Takens-based approaches are not a reliable option for studying many real-world systems. The reduced order modeling approach, as in the first part of this paper, attempts to tackle this fundamental limitation, by exploiting the multiscale structure of real-world dynamical systems and focusing on coarse-grained effective dynamics. The success of GFDT in this context hinges on two key conditions: a clear timescale separation between the resolved slow modes and unresolved fast modes, and a perturbation that projects primarily onto these slow modes. When these conditions hold, as is common for large-scale climate forcings, the cumulative effect of unresolved variables can be modeled as stochastic noise, and response theory can be effectively applied to the coarse-grained dynamics \citep{Hasselmann1976,Penland89,majda_fdt,LucariniChekroun2023,Lacorata,lucarini2025interpretable,falasca2025neuralmodelsmultiscalesystems}. The success of this proposed method in real-world systems depends critically on the choice of the \textit{proper} variables used to study the phenomena of interest; we refer the reader to the contribution in \cite{falasca2025neuralmodelsmultiscalesystems} for more details. While our results focused on data-driven strategies applied to pre-defined ROMs, future work will extend these ideas to the more challenging scenario of inferring responses directly from partially observed data of high-dimensional systems.
    \item \textit{Absence of scale separation.} Finally, it is possible to envision conditions where we want to study the response to perturbations coupling with all modes. The feasibility of such approach will depend on observing of the full state vector, or at least all the relevant proper variables. Our analysis of the 2D Navier-Stokes system, with its localized perturbation coupling to all spatial scales, exemplifies such a case. Here, a ROM is inadequate, and a direct estimation of the score function on the high-dimensional state space is required. Our use of a U-Net architecture demonstrates that this is a viable, albeit more computationally demanding, alternative.
\end{itemize}

\section{Conclusions}
\label{sec:conclusions}

In this work, we have introduced a framework for constructing higher-order response functions in nonlinear stochastic systems by combining the Generalized Fluctuation-Dissipation Theorem (GFDT) with score-based generative modeling. Our method circumvents the limitations of traditional Gaussian approximations by estimating the score function directly from data, allowing for accurate prediction of moment responses and perturbed probability density functions (PDFs) in non-Gaussian regimes.\\

{We have validated our approach on a hierarchy of systems. For systems admitting a low-dimensional effective description, we used the KGMM algorithm to estimate the score function for three prototypical models: a scalar stochastic model, a slow-fast triad model for ENSO dynamics, and a six-dimensional stochastic barotropic model. In all cases, the GFDT framework equipped with KGMM-estimated scores accurately captured higher-order statistical responses, significantly outperforming the linear Gaussian approximation. Furthermore, we demonstrated the framework's versatility by extending it to a high-dimensional 2D Navier-Stokes system, where a reduced-order model is not suitable due to a localized perturbation. By employing a U-Net to learn the score function directly in the high-dimensional space, we again achieved accurate response predictions that surpassed the Gaussian approximation.} \\

{These results showcase the power and flexibility of combining GFDT with data-driven score estimation. The choice between a clustering-based method like KGMM for low-dimensional ROMs and a deep learning approach like a U-Net for full high-dimensional systems provides a versatile toolkit for tackling a wide range of problems.} Future work will focus on extending this approach to larger-scale systems with partial observations and on {systematically comparing the trade-offs between ROM-based and full-system response predictions to develop robust best practices for real-world applications.}

\section*{Acknowledgments}
{This work acknowledges support by Schmidt Sciences, LLC, through the Bringing Computation to the Climate Challenge, an MIT Climate Grand Challenge Project. We would also like to thank the community at Altamira Collaboratory for providing a platform for the discussion of the ideas herein, Aeolus Labs and the Geophysical Fluid Dynamics Program program where part of this work was completed. The authors also thank Valerio Lucarini, Marco Baldovin, and an anonymous reviewer for useful comments that helped greatly improve the presentation and quality of the present manuscript.}

\appendix
\clearpage
\section*{Appendix}
\addcontentsline{toc}{section}{Appendix}

\input{SM_appendix}

\bibliographystyle{plainnat}
\bibliography{references}

\end{document}

%% file: SM_appendix.tex
\section{Reconstruction of the Perturbed Steady-State Distribution via the Maximum Entropy Principle}\label{sm:maxent}

In this section, we describe how the perturbed steady-state probability distributions can be reconstructed from the moment responses obtained through the GFDT framework. This methodology complements the main analysis presented in Section 3 of the manuscript by providing a way to visualize the full distributional response rather than just individual moments.

The GFDT framework allows us to estimate the perturbed steady-state probability density function \(\rho_1(\bm{x})\) without directly simulating the perturbed system. Theoretically, this distribution satisfies the stationary Fokker–Planck equation associated with the perturbed dynamics:
\begin{equation}
\mathcal{L}_0 \rho_1(\bm{x}) + \mathcal{L}_1 \rho_S(\bm{x}) = 0,
\end{equation}
where \(\mathcal{L}_0\) is the unperturbed Fokker–Planck operator and \(\mathcal{L}_1\) encodes the effect of the small perturbation, as defined in Section 2.1.

Rather than solving this partial differential equation directly, we infer the change in the distribution from its effect on a finite set of observables. In particular, we apply GFDT to observables corresponding to the first \(N\) powers of the system state \(\bm{x}\), such as \(\bm{x}\), \(\bm{x} \bm{x}^\top\), \(\bm{x}^{\otimes 3}\), and so on. This allows us to estimate the perturbation-induced variation in the first \(N\) moments of the steady-state distribution. For instance, the change in the first two moments is given by:
\begin{equation}
\begin{split}
\delta\bm{m}_1 &= \mathbb{E}_1[\bm{x}] - \mathbb{E}_0[\bm{x}] = \mathbb{E}_1[\bm{x}] - \bm{\mu} = \langle \delta \bm{x} \rangle, \\
\delta\bm{m}_2 &= \mathbb{E}_1[(\bm{x} - \mathbb{E}_1[\bm{x}])(\bm{x} - \mathbb{E}_1[\bm{x}])^\top] - \mathbb{E}_0[(\bm{x} - \bm{\mu})(\bm{x} - \bm{\mu})^\top] \\
&\approx \mathbb{E}_1[(\bm{x} - \bm{\mu})(\bm{x} - \bm{\mu})^\top] - \mathbb{E}_0[(\bm{x} - \bm{\mu})(\bm{x} - \bm{\mu})^\top] = \langle \delta[(\bm{x} - \bm{\mu})(\bm{x} - \bm{\mu})^\top] \rangle.
\label{eq:moments}
\end{split}
\end{equation}

These quantities provide indirect but crucial information about how the perturbed steady-state distribution \(\rho_1(\bm{x})\) differs from the unperturbed one \(\rho_S(\bm{x})\), even in the absence of direct samples from the perturbed dynamics.

To reconstruct the perturbed distribution \(\rho_1(\bm{x})\) consistent with the estimated moments, we invoke the maximum entropy principle. Among all candidate distributions satisfying the known moment constraints, this principle selects the one with maximal Shannon entropy, ensuring the most unbiased estimate compatible with the available information \citep{mead1984maximum}.

Under this principle, the perturbed probability density function takes the exponential form:
\begin{equation}
\rho_1(\bm{x}) = \exp\!\left( \sum_{i=0}^{N} \bm{\lambda}_i \cdot \bm{\phi}_i(\bm{x}) \right),
\end{equation}
where \(\bm{\phi}_i(\bm{x})\) denotes the basis functions associated with the \(i\)-th moment (e.g., monomials, tensor products), and \(\bm{\lambda}_i\) are Lagrange multipliers enforcing the moment constraints. These coefficients are obtained by solving the system:
\begin{equation}
\int_{\bm{\Omega}} \bm{\phi}_i(\bm{x})\,\rho_1(\bm{x})\,\mathrm{d}\bm{x} = \bm{m}_i, \quad \text{for } i = 0, 1, \ldots, N,
\end{equation}
where \(\bm{m}_0 = 1\) ensures normalization, and \(\bm{m}_1, \ldots, \bm{m}_{N}\) are the perturbed moments computed via GFDT. In this work, we restrict the basis functions \(\bm{\phi}_i(\bm{x})\) to be multivariate polynomials, corresponding to the moment tensors of increasing order defined in Eq. \eqref{eq:moments}.

This system of nonlinear integral equations generally lacks a closed-form solution and is solved numerically, for instance via iterative root-finding or variational methods. In our implementation, we first generate multiple candidate starting points based on heuristic estimates of key statistics (mean, variance, skewness, etc.) derived from the target moments. These candidates are then optimized in parallel using the Nelder–Mead method, complemented by preliminary LBFGS refinement to minimize the discrepancy between computed and desired moments. The best candidate undergoes a regularized Newton–Raphson refinement—where the Jacobian of the moment equations is computed using adaptive quadrature—to ensure rapid convergence toward the solution. Finally, a polished LBFGS optimization step is applied to further reduce the error. Once the multipliers \(\bm{\lambda}_i\) are known, the resulting maximum entropy distribution provides an approximation of the perturbed steady-state PDF that incorporates the estimated changes in mean, variance, and higher-order moments.\\

In this study, the Maximum Entropy Principle is used primarily as a proof of concept, applied to the scalar stochastic model discussed in Section 3.1 of the manuscript. This application allows us to validate the accuracy of our GFDT predictions by comparing the reconstructed perturbed distributions against ground truth distributions obtained analytically.\\

\begin{figure}[H]
    \centering
    \includegraphics[width=\textwidth]{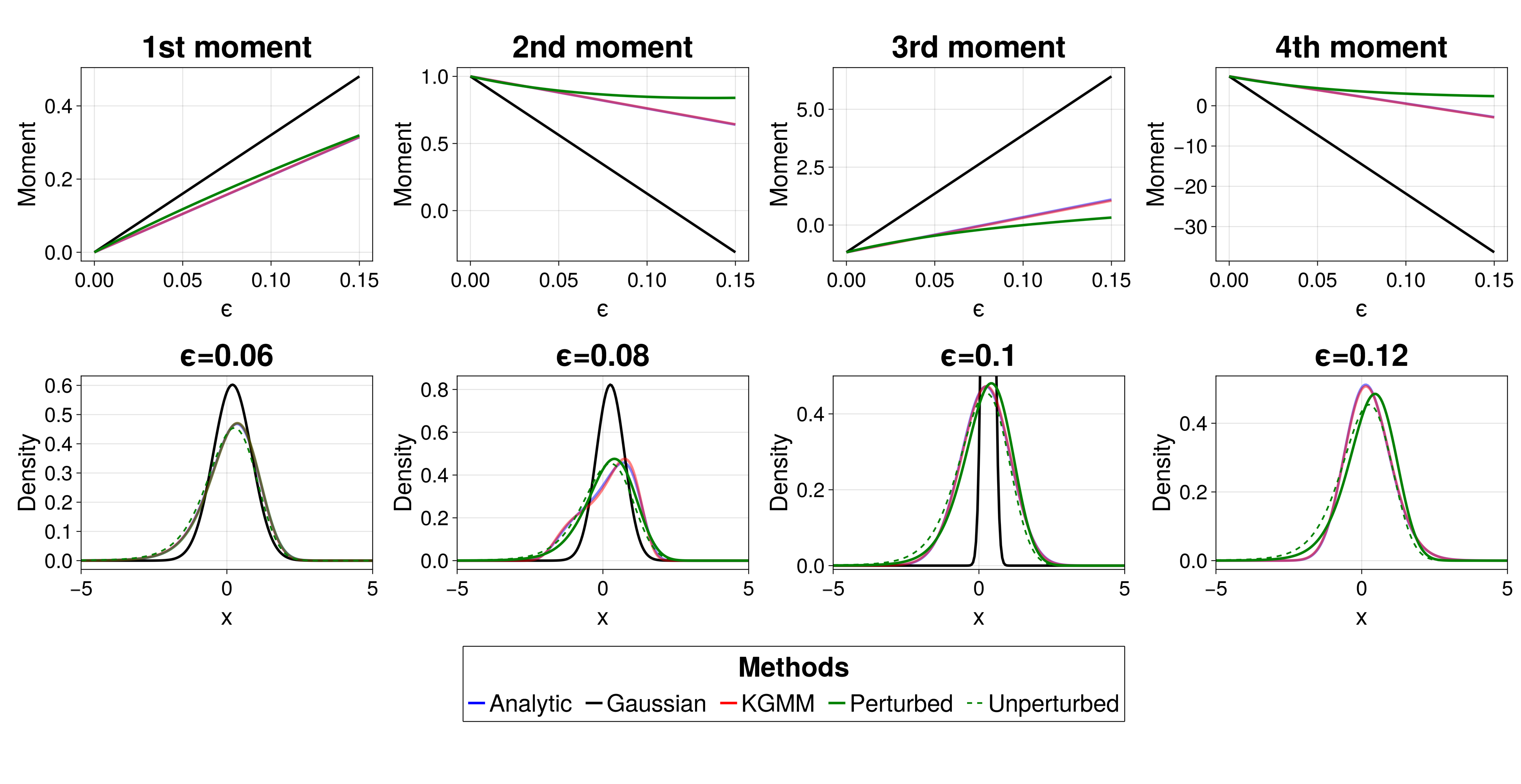}
    \caption{Top row: First four statistical moments—mean, variance, skewness, and kurtosis—as a function of the perturbation amplitude $\epsilon$, obtained using the analytic expression of the PDF, and GFDT with the analytic (blue), linear (black) and KGMM (red) score function. Bottom row: Comparison between the true perturbed PDF (green) and the PDFs reconstructed via the maximum entropy method and GFDT for $\epsilon = 0.06$, $0.08$, $0.10$, and $0.12$. We compare three approaches: reconstruction using the analytic score (blue), the KGMM-estimated score (red), and a linear (Gaussian) approximation (black).}
    \label{fig:reduced_moments}
\end{figure}

For this demonstration, we introduced a constant perturbation in the forcing term of the scalar model, shifting $F \rightarrow F + \epsilon$, and studied the response of the first four moments as a function of the perturbation amplitude $\epsilon$. Figure~\ref{fig:reduced_moments} shows both the moment responses and the reconstructed PDFs for different values of $\epsilon$. The results confirm that up to $\epsilon = 0.06$, the responses predicted using both the analytic score function and the KGMM-estimated score via GFDT match the true moment variations very closely, validating that the linear-response regime holds within this range. For larger perturbations, increasing discrepancies appear, highlighting the breakdown of the linear response assumption.

The reconstructed perturbed PDFs shown in the bottom row of Figure~\ref{fig:reduced_moments} were estimated using the maximum entropy principle, based on the first two moments for the Gaussian approximation and the first four moments for the analytic and KGMM-based approaches (for $\epsilon=0.06, 0.08$), and the first three moments for $\epsilon=0.10, 0.12$. For these larger values of $\epsilon$, including more moments in the maximum entropy reconstruction leads to convergence towards a multimodal distribution, which would be impossible to capture with a Gaussian approximation. Indeed, for $\epsilon=0.12$, the algorithm used to reconstruct the perturbed PDF from the Gaussian approximation failed to converge.

While this approach worked well in the low-dimensional setting of the scalar model, its application to the higher-dimensional models presented in this paper would be computationally demanding and is beyond the scope of the current work. Nevertheless, the results from this simple case clearly demonstrate how the KGMM-based GFDT approach outperforms traditional linear approximations in predicting both the individual moments and the overall shape of perturbed probability distributions.

\section{Demonstration of Probability Conservation}\label{sm:prob_cons}

For the perturbed density $\rho_S + \delta\rho$ to remain a valid probability distribution, the total probability must be conserved. This requires the integral of the density variation over the entire state space to be zero at all times. This consistency constraint, $\int \delta\rho(\mathbf{x},t)\,d\mathbf{x} = 0$, is satisfied by Equation (5) of the main text. To prove this, we demonstrate that its time derivative is zero and that it holds true at the initial time, $t=0$.

We begin with the linearized Fokker-Planck equation, which is Equation (5) from the main text:
\begin{equation}
\label{eq:linearized_FP}
\frac{\partial \delta\rho(\bm{x},t)}{\partial t} = \mathcal{L}_0\,\delta\rho(\bm{x},t) + f(t)\,\mathcal{L}_1\,\rho_S(\bm{x})
\end{equation}
Integrating both sides over the entire state space $\mathbb{R}^n$ yields:
\begin{equation}
\label{eq:integrated_FP}
\int \frac{\partial \delta\rho(\bm{x},t)}{\partial t} d\bm{x} = \int \mathcal{L}_0\,\delta\rho(\bm{x},t) d\bm{x} + f(t) \int \mathcal{L}_1\,\rho_S(\bm{x}) d\bm{x}
\end{equation}
We can swap the order of integration and differentiation on the left-hand side, leading to:
\begin{equation}
\label{eq:time_derivative_integral}
\frac{d}{dt} \int \delta\rho(\bm{x},t) d\bm{x} = \int \mathcal{L}_0\,\delta\rho(\bm{x},t) d\bm{x} + f(t) \int \mathcal{L}_1\,\rho_S(\bm{x}) d\bm{x}
\end{equation}

Next, we evaluate the two integrals on the right-hand side. The Fokker-Planck operator $\mathcal{L}_0$ is defined as $\mathcal{L}_0\,\rho = -\nabla\cdot\big(\bm{F}\,\rho\big) + \frac{1}{2}\nabla\nabla^\top : \big(\bm{D}\,\rho\big)$. Its integral over the whole domain is:
\begin{equation}
\int \mathcal{L}_0\,\delta\rho(\bm{x},t) d\bm{x} = \int \left( -\nabla\cdot\big(\bm{F}(\bm{x})\,\delta\rho(\bm{x},t)\big) + \frac{1}{2}\nabla\nabla^\top : \big(\bm{D}\,\delta\rho(\bm{x},t)\big) \right) d\bm{x}
\end{equation}
By applying the divergence theorem, the integral of the divergence terms can be converted into surface integrals at the boundary (at infinity). Assuming no-flux boundary conditions, meaning the probability current is zero at infinity, these surface integrals vanish. Thus, the integral of the Fokker-Planck operator over the entire domain is zero:
\begin{equation}
\label{eq:L0_integral_zero}
\int \mathcal{L}_0\,\delta\rho(\bm{x},t) d\bm{x} = 0
\end{equation}

Similarly, for the perturbation operator $\mathcal{L}_1$, defined as $\mathcal{L}_1\,\rho = -\nabla\cdot\big(\bm{u}(\bm{x})\,\rho\big)$, its integral is:
\begin{equation}
\int \mathcal{L}_1\,\rho_S(\bm{x}) d\bm{x} = \int -\nabla\cdot\big(\bm{u}(\bm{x})\,\rho_S(\bm{x})\big) d\bm{x}
\end{equation}
Again, utilizing the divergence theorem and assuming the term $\bm{u}(\bm{x})\rho_S(\bm{x})$ vanishes at infinity (a reasonable assumption as the steady-state density $\rho_S$ must vanish at infinity), this integral is also zero:
\begin{equation}
\label{eq:L1_integral_zero}
\int \mathcal{L}_1\,\rho_S(\bm{x}) d\bm{x} = 0
\end{equation}

Substituting Equations \eqref{eq:L0_integral_zero} and \eqref{eq:L1_integral_zero} back into Equation \eqref{eq:time_derivative_integral}, we obtain:
\begin{equation}
\frac{d}{dt} \int \delta\rho(\bm{x},t) d\bm{x} = 0 + f(t) \cdot 0 = 0
\end{equation}
This crucial result demonstrates that the total integrated density perturbation, $\int\delta\rho(\bm{x},t)d\bm{x}$, does not change in time.

Furthermore, given that the system is unperturbed at $t=0$, the density perturbation at that time is zero everywhere, $\delta\rho(\bm{x},0) = 0$. Consequently, its integral at $t=0$ is also zero:
\begin{equation}
\int\delta\rho(\bm{x},0)d\bm{x} = 0
\end{equation}
Since the value of the integral is zero at $t=0$ and its time derivative is always zero, it must remain zero for all $t \ge 0$.

\section{Derivation of the First-Order Response Formula}\label{sm:first_order}

In this section, we provide a derivation of the first-order correction to the expectation value of an observable, as presented in Eq. 7 of the main text. Our starting point is the linearized Fokker--Planck equation (Eq. 5), which describes the evolution of the probability density perturbation, $\delta\rho(\bm{x},t)$, in response to a small, time-dependent perturbation.

The linearized dynamics for $\delta\rho(\bm{x},t)$ are given by:
\begin{equation}
\frac{\partial \delta\rho(\bm{x},t)}{\partial t} = \mathcal{L}_0\,\delta\rho(\bm{x},t) + f(t)\,\mathcal{L}_1\,\rho_S(\bm{x}),
\end{equation}
where we assume the system was in its steady state for $t < 0$, implying $\delta\rho(\bm{x},0) = 0$. This linear inhomogeneous differential equation can be formally solved by integrating with respect to time, which yields an expression for the density perturbation:
\begin{equation}
\delta\rho(\bm{x},t) = \int_0^t e^{\mathcal{L}_0(t-t')} f(t') \mathcal{L}_1\,\rho_S(\bm{x}) \mathrm{d}t'.
\end{equation}
Here, the operator $e^{\mathcal{L}_0(t-t')}$ is the propagator corresponding to the unperturbed Fokker--Planck operator $\mathcal{L}_0$, which evolves the system from time $t'$ to $t$.

The first-order change in the expectation value of an observable $A(\bm{x})$, denoted $\langle \delta A(t) \rangle$, is obtained by integrating $A(\bm{x})$ with this density perturbation:
\begin{equation}
\langle \delta A(t) \rangle = \int A(\bm{x}) \, \delta\rho(\bm{x},t) \, \mathrm{d}\bm{x}.
\end{equation}
Substituting the formal solution for $\delta\rho(\bm{x},t)$ from Eq. (S2) and swapping the order of integration gives:
\begin{equation}
\langle \delta A(t) \rangle = \int_0^t f(t') \left( \int A(\bm{x}) \, e^{\mathcal{L}_0(t-t')} (\mathcal{L}_1\,\rho_S(\bm{x})) \, \mathrm{d}\bm{x} \right) \mathrm{d}t'.
\label{eq:delta_A}
\end{equation}
By using the property of the adjoint of the Fokker--Planck operator, $\mathcal{L}_0^\dagger$, which dictates the time evolution of observables in the unperturbed system, we can transfer the action of the propagator $e^{\mathcal{L}_0(t-t')}$ from the density onto the observable $A(\bm{x})$. The term $e^{\mathcal{L}_0^\dagger(t-t')}A(\bm{x})$ corresponds to the observable $A$ at time $t$ for a trajectory that was at state $\bm{x}$ at time $t'$, which we denote as $A(\bm{x}(t))$. The inner integral in equation \eqref{eq:delta_A} then becomes:
\begin{equation}
\int \left(e^{\mathcal{L}_0^\dagger(t-t')} A(\bm{x})\right) \, (\mathcal{L}_1\,\rho_S(\bm{x})) \, \mathrm{d}\bm{x}.
\end{equation}
Using the definition of the conjugate observable $B(\bm{x}) = (\mathcal{L}_1\,\rho_S(\bm{x}))/\rho_S(\bm{x})$ from Eq. 8 in the main text, we can rewrite Eq. (S5) as:
\begin{equation}
\int A(\bm{x}(t)) \, B(\bm{x}(t')) \, \rho_S(\bm{x}(t')) \, \mathrm{d}\bm{x}(t').
\end{equation}
This expression is precisely the definition of the two-time correlation function between observable $A$ at time $t$ and observable $B$ at time $t'$, calculated in the unperturbed steady state. We denote this average as $\langle A(\bm{x}(t))\,B(\bm{x}(t'))\rangle_0$. Substituting this back into equation \eqref{eq:delta_A} yields the first form of the response formula:
\begin{equation}
\langle \delta A(t) \rangle = \int_0^t \! f(t')\,\big\langle A(\bm{x}(t))\,B(\bm{x}(t'))\big\rangle_0 \mathrm{d}t'.
\end{equation}
Finally, because the unperturbed system is in a steady state, its statistics are stationary in time. This implies that the correlation function depends only on the time lag $\tau = t - t'$, i.e., $\langle A(\bm{x}(t))\,B(\bm{x}(t'))\rangle_0 = \langle A(\bm{x}(t-t'))\,B(\bm{x}(0))\rangle_0$. This latter quantity is defined as the linear response function, $\bm{R}(t-t')$. This leads to the final convolution integral form presented in Eq. 7 of the main text:
\begin{equation}
\langle \delta A(t) \rangle = \int_0^t \bm{R}(t - t')\,f(t')\,\mathrm{d}t'.
\end{equation}

\section{Response of Central Moments to a Constant Perturbation}\label{sm:central_moments}

In this section, we analyze the instantaneous response ($t=0$) of the statistical central moments of a system to a small, constant external perturbation. This provides insight into how the probability distribution initially shifts. We define the unperturbed mean of the i-th component as $\mu_i = \langle u_i \rangle$.

We consider a constant perturbation field, denoted by the vector $\vec{\delta}$. According to the Generalized Fluctuation-Dissipation Theorem (GFDT), the response of an observable $A(\vec{u})$ at $t=0$ is given by its correlation with a conjugate observable, $B(\vec{u}) = -\vec{\delta} \cdot \vec{s}(\vec{u})$, where $\vec{s}(\vec{u}) = \nabla \ln \rho_S(\vec{u})$ is the score function of the unperturbed steady-state distribution $\rho_S$.

The observable of interest here is the n-th central moment of the i-th component of the state vector, $A(\vec{u}) = (u_i - \mu_i)^n$. The response of this moment, which we denote as $R_{\text{cen}, n}(i, \vec{\delta})$, is:
\begin{equation}
    R_{\text{cen}, n}(i, \vec{\delta}) = \langle (u_i - \mu_i)^n B(\vec{u}) \rangle = \langle (u_i - \mu_i)^n (-\vec{\delta} \cdot \vec{s}(\vec{u})) \rangle = -\sum_{j} \delta_j \langle (u_i - \mu_i)^n s_j(\vec{u}) \rangle
\end{equation}

To derive a general expression, we evaluate the expectation term $\langle (u_i - \mu_i)^n s_j(\vec{u}) \rangle$ using integration by parts, assuming that the probability density $\rho_S(\vec{u})$ vanishes at the boundaries of the domain $\Omega$:
\begin{align}
    \langle (u_i - \mu_i)^n s_j(\vec{u}) \rangle &= \int_{\Omega} (u_i - \mu_i)^n s_j(\vec{u}) \rho_S(\vec{u}) \,d\vec{u} = \int_{\Omega} (u_i - \mu_i)^n \frac{\partial\rho_S(\vec{u})}{\partial u_j} \,d\vec{u} \nonumber \\
    &= - \int_{\Omega} \frac{\partial (u_i - \mu_i)^n}{\partial u_j} \rho_S(\vec{u}) \,d\vec{u}
\end{align}
Since $\frac{\partial (u_i - \mu_i)^n}{\partial u_j} = n (u_i - \mu_i)^{n-1} \delta_{ij}$, where $\delta_{ij}$ is the Kronecker delta, the expression simplifies to:
\begin{equation}
    \langle (u_i - \mu_i)^n s_j(\vec{u}) \rangle = - \int_{\Omega} n (u_i - \mu_i)^{n-1} \delta_{ij} \rho_S(\vec{u}) \,d\vec{u} = -n \delta_{ij} \langle (u_i - \mu_i)^{n-1} \rangle
\end{equation}
Substituting this result back into the equation for the response $R_{\text{cen}, n}$, we obtain the general expression for the instantaneous response of the n-th central moment:
\begin{equation}
    R_{\text{cen}, n}(i, \vec{\delta}) = -\sum_{j} \delta_j (-n \delta_{ij} \langle (u_i - \mu_i)^{n-1} \rangle) = n \delta_i \langle (u_i - \mu_i)^{n-1} \rangle
\end{equation}
We now examine this general expression for the first and second central moments.

\paragraph{First Central Moment (n=1)}
The observable is $(u_i - \mu_i)$. Its response is:
\begin{equation}
    R_{\text{cen}, 1}(i, \vec{\delta}) = 1 \cdot \delta_i \langle (u_i - \mu_i)^0 \rangle = \delta_i \langle 1 \rangle = \delta_i
\end{equation}
This represents the change in the expected value of $(u_i - \mu_i)$. Since the unperturbed expectation $\langle u_i - \mu_i \rangle$ is zero, the new expectation is simply $\delta_i$. This is consistent with the mean of $u_i$ shifting from $\mu_i$ to $\mu_i + \delta_i$. The response depends on the perturbation component $\delta_i$ but is independent of the overall perturbation magnitude $|\vec{\delta}|$ and the system's state $\vec{u}$.

\paragraph{Second Central Moment (n=2)}
The observable is $(u_i - \mu_i)^2$, which is the variance. Its response is:
\begin{equation}
    R_{\text{cen}, 2}(i, \vec{\delta}) = 2 \cdot \delta_i \langle (u_i - \mu_i)^1 \rangle = 2 \delta_i \cdot 0 = 0
\end{equation}
The linear response of the second central moment (variance) is zero. This means that to the first order in the perturbation $\vec{\delta}$, there is no change in the system's variance. Any change in variance is a higher-order effect (e.g., proportional to $\delta^2$).

\paragraph{Note on Raw vs. Central Moments}
It is important to distinguish the response of central moments from that of raw moments. The property that the second moment's response depends on the perturbation $\vec{\delta}$ but not on the state $\vec{u}$ applies specifically to the second \textit{raw} moment ($A(\vec{u}) = u_i^2$). For completeness, the response of the second raw moment is $R_{\text{raw}, 2}(i, \vec{\delta}) = 2\delta_i \langle u_i \rangle$. This response depends on the perturbation component $\delta_i$ and the unperturbed mean $\langle u_i \rangle$ (a constant), but not on the instantaneous state variable.

\section{Parameter Tables}\label{sm:params}

This section provides the parameter values used for the three reduced-order models analyzed in the main text.

\begin{table}[H]
\centering
\caption{Parameters of the stochastic scalar model.}
\label{tab:triad_params}
\begin{tabular}{lc}
\toprule
$a$       & $-0.0222$ \\
$b$       & $-0.2$ \\
$c$       & $0.0494$ \\
$F$       & $0.6$ \\
$\sigma$  & $0.7071$ \\
\bottomrule
\end{tabular}
\end{table}

\begin{table}[H]
\centering
\begin{tabular}{lc}
\toprule
$d_{u}$ & $0.2$ \\
$\omega$ & $0.4$ \\
$d_{\tau}$ & $2.0$ \\
$\sigma_{u1}=\sigma_{u2}$ & $0.3$ \\
$\sigma_{\tau}(u_1)$ & $1.5(\tanh(u_1)+1)$ \\
\bottomrule
\end{tabular}
\caption{Parameters of the slow-fast triad model.}
\label{tab:enso_params}
\end{table}

\begin{table}[h]
\centering
\caption{Model coefficients for the six-dimensional stochastic barotropic system.}
\label{tab:params}
\begin{tabular}{lll}
\toprule
\textbf{Parameter} & \textbf{Value} & \textbf{Description} \\
\midrule
$\alpha_1$         & 0.86322463 & Nonlinear advection (mode 1) \\
$\alpha_2$         & 0.81394451 & Nonlinear advection (mode 2) \\
$\beta_1$          & 0.89887640 & Coriolis effect (mode 1) \\
$\beta_2$          & 0.48780488 & Coriolis effect (mode 2) \\
$\delta_1$         & 1.38115941 & Triad interaction (mode 1) \\
$\delta_2$         & $-0.12882575$ & Triad interaction (mode 2) \\
$\tilde{\gamma}_1$ & 0.19206748 & Orographic forcing (mode 1) \\
$\tilde{\gamma}_2$ & 0.07682699 & Orographic forcing (mode 2) \\
$\gamma_1$         & 0.05395154 & Orographic damping (mode 1) \\
$\gamma_2$         & 0.04684573 & Orographic damping (mode 2) \\
$\varepsilon$      & 1.44050611 & Wave-wave interaction \\
$C$                & 0.1        & Newtonian relaxation rate \\
$x_1^*$            & 0.95       & Zonal background forcing (mode 1) \\
$x_4^*$            & $-0.76095$ & Zonal background forcing (mode 2) \\
$\sigma$           & 0.01       & Noise amplitude \\
\bottomrule
\end{tabular}
\end{table}

\section{Data-Driven Score Function Estimation}\label{sm:score_overview}

The practical application of the Generalized Fluctuation-Dissipation Theorem (GFDT) hinges on the ability to accurately estimate the score function, $\nabla \ln \rho_S(\bm{x})$, from data. In this section, we provide the technical details for the two complementary, data-driven methods used in the main text: the KGMM algorithm for low-dimensional systems and the U-Net-based denoising score matching approach for high-dimensional systems.

\subsection{Score Function Estimation via KGMM}\label{sm:kgmm}

The application of the Generalized Fluctuation-Dissipation Theorem (GFDT) requires knowledge of the score function $\nabla \ln \rho_S(x)$, i.e., the gradient of the logarithm of the system's steady-state distribution. For the vast majority of systems of interest, this quantity cannot be obtained analytically and must be inferred from data. To this end, a hybrid statistical-learning method called KGMM (K-means Gaussian Mixture Modeling) has been recently proposed \cite{giorgini2025kgmm} for accurate and efficient score function estimation in systems with low-dimensional \textit{effective} dynamics.

The KGMM method is based on the observation that a probability density can be approximated as a Gaussian Mixture Model (GMM):
\begin{equation}
p(\bm{x}) = \frac{1}{N} \sum_{i=1}^{N} \mathcal{N}(\bm{x} \mid \bm{\mu}_i, \sigma_G^2 \bm{I}),
\end{equation}
where the $\bm{\mu}_i$ are the $N$ data samples drawn from the steady-state distribution $\rho_S(\bm{x})$, and $\sigma_G^2$ is the (isotropic) covariance amplitude of the Gaussian kernels. The corresponding score function reads
\begin{equation}
\nabla \ln p(\bm{x}) = -\frac{1}{\sigma_G^2} \sum_{i=1}^{N} \frac{\mathcal{N}(\bm{x} \mid \bm{\mu}_i, \sigma_G^2 \bm{I})(\bm{x} - \bm{\mu}_i)}{p(\bm{x})}.
\label{eq:gmm_score}
\end{equation}
While Eq.~\eqref{eq:gmm_score} provides a direct expression for the score, it becomes numerically unstable for small $\sigma_G$, as the density and its derivative become highly sensitive to local data fluctuations.

KGMM circumvents this issue by exploiting a probabilistic identity: define $\bm{x} = \bm{\mu} + \sigma_G \bm{z}$ with $\bm{z} \sim \mathcal{N}(0, \bm{I})$. \cite{giorgini2025kgmm} derived the following relation in the continuous-data limit:
\begin{equation}
\nabla \ln p(\bm{x}) = -\frac{1}{\sigma_G^2} \mathbb{E}[\bm{z} \mid \bm{x}],
\label{eq:score_identity}
\end{equation}
i.e., the score function is the conditional expectation of the kernel displacements $\bm{z}$, rescaled by $\sigma_G^2$.

This identity is computed empirically using the following steps:
\begin{enumerate}
    \item Generate perturbed samples $\bm{x}_i = \bm{\mu}_i + \sigma_G \bm{z}_i$, where $\bm{\mu}_i$ are the original data points and $\bm{z}_i \sim \mathcal{N}(0, \bm{I})$.
    \item Partition the perturbed samples $\{\bm{x}_i\}$ into $N_C$ clusters $\{\Omega_j\}$ using a modified bisecting K-means clustering \citep{souza2024modified}.
    \item For each cluster $\Omega_j$ with centroid $\bm{C}_j$, compute the conditional expectation
    \begin{equation}
    \mathbb{E}[\bm{z} \mid \bm{x} \in \Omega_j] \approx \frac{1}{|\Omega_j|} \sum_{i: \bm{x}_i \in \Omega_j} \bm{z}_i.
    \end{equation}
    \item Estimate the score function at the centroid $\bm{C}_j$ as
    \begin{equation}
    \nabla \ln \rho_S(\bm{C}_j) \approx -\frac{1}{\sigma_G^2} \mathbb{E}[\bm{z} \mid \bm{x} \in \Omega_j].
    \end{equation}
    \item Fit a neural network to interpolate the discrete estimates $\{(\bm{C}_j, \nabla \ln \rho_S(\bm{C}_j))\}$ over the full domain.
\end{enumerate}

The number of clusters \(N_C\) must therefore be chosen carefully to balance the trade-off between resolution and noise. Empirically, a useful, approximate scaling relation is
\begin{equation}
N_C \propto \sigma_G^{-d},
\end{equation}
where \(d\) is the effective dimensionality of the dataset and \(\sigma_G\) is the kernel width. This scaling ensures that clusters remain small enough to capture local gradient structure while still containing enough points for robust averaging. It is important to note that using a single, global kernel width $\sigma_G$ imposes a uniform resolution across the entire state space. This simplification may lead to the loss of fine-grained local features in regions where the score function has high curvature. Future work could explore adaptive approaches where $\sigma_G$ is varied based on local data density or other indicators to achieve a more uniform effective resolution.

The choice of \(\sigma_G\) plays a central role in the KGMM algorithm. Small values of \(\sigma_G\) yield estimates of the score function that are closer to the one associated with the true steady-state distribution, as the perturbation introduced by the convolution kernel becomes negligible. However, this comes at the cost of increased statistical noise, since the displacements become more sensitive to sample variability. Conversely, larger values of \(\sigma_G\) smooth out the fluctuations, leading to more stable estimates but of a score function associated with a more strongly perturbed distribution. The optimal value of \(\sigma_G\) thus balances these competing effects---reducing bias while maintaining statistical reliability. In practice, $\sigma_G$ is a hyperparameter determined empirically. For all reduced-order models presented in this work, we found that a value of $\sigma_G=0.05$ provided a robust balance, yielding accurate score estimates. This choice is consistent with practical guidelines from related literature \citep{bischoff2024unpaired}, which suggest that for good performance the value of $\sigma_G$ should be larger than $10^{-2}$, a guideline that our work confirms.

For the applications presented in this work, we used $10^6$ uncorrelated samples from the steady-state distribution to train the KGMM algorithm. However, the method works well also with much fewer data points; in the Supplementary Material, we present the same results using $10^4$ uncorrelated points instead, demonstrating the robustness of the approach to dataset size.

To interpolate the discrete score estimates $(\bm{C}_j, \nabla \ln \rho_S(\bm{C}_j))$, we train a fully connected feedforward neural network, also known as a Multi-Layer Perceptron (MLP). This choice is motivated by the universal approximation theorem, which states that an MLP can approximate any continuous function to arbitrary precision, making it a suitable tool for learning the smooth score function \citep{goodfellow2016deep}. For this regression task on low-to-moderate dimensional data, an MLP provides a robust and computationally efficient solution. We employ the Swish activation function between layers, as its smoothness and non-monotonicity have been shown to improve performance in deep networks \citep{ramachandran2017searching}. The models and training parameters are:
\begin{itemize}
  \item \textbf{Scalar model:} Hidden layers of 50 and 25 neurons; batch size 32; 2000 epochs.
  \item \textbf{Slow–fast triad model:} Hidden layers of 100 and 50 neurons; batch size 32; 200 epochs.
  \item \textbf{Barotropic model:} Hidden layers of 128 and 64 neurons; batch size 128; 300 epochs.
\end{itemize}
These specific architectures and training parameters were determined through standard hyperparameter tuning. The number of layers and neurons for each model was chosen to provide sufficient capacity to capture the complexity of the respective score function without overfitting the discrete training data. The final parameters were chosen to minimize the training loss, i.e., the mean-squared error between the network's output and the discrete score estimates.
All networks are trained using the Adam optimizer with mean-squared-error loss on the predicted score \citep{kingma2014adam}.

\subsection{Score Function Estimation via U-Net}\label{app:UNet}

When reduced-order models (ROMs) cannot be effectively applied—either because there is no clear timescale separation between fast and slow modes, or because the perturbation couples with all modes of the high-dimensional system—the application of the clustering algorithm becomes computationally infeasible in high-dimensional spaces. In such cases, we must directly estimate the score function for the full system without dimensional reduction.

For high-dimensional systems, we train a neural network to learn the score identity from Eq.~\ref{eq:score_identity} directly using a denoising score matching approach. This method circumvents the need for explicit clustering by leveraging the relationship between the score function and conditional expectations of noise perturbations. We employ the same U-Net architecture used in \citep{giorgini_response_theory}, which is specifically designed for learning score functions of spatially-extended systems and consists of an encoder-decoder structure with skip connections that preserve spatial information across different resolution scales.

The U-Net architecture comprises the following components:
\begin{itemize}
    \item \textbf{Lifting layer:} Preserves spatial dimensions while increasing channel count from 1 to 32.
    \item \textbf{Downsampling layers:} Three layers reducing spatial size by a factor of 2 each, with channel progression 32→64→128→256.
    \item \textbf{Residual blocks:} Eight residual blocks maintaining spatial and channel dimensions at the bottleneck.
    \item \textbf{Upsampling layers:} Three layers increasing spatial size by a factor of 2 each with channel reduction 256→128→64→32.
    \item \textbf{Projection layer:} Final layer outputting a single channel result.
\end{itemize}

All convolutions respect the periodicity of the system using periodic padding.

The network is trained to minimize the denoising score matching loss:
\begin{equation}
\mathcal{L}(\theta) = \mathbb{E}\left[\sigma_G^2 \left\|\mathbf{s}_\theta(\bm{\mu} + \sigma_G \mathbf{z}) + \frac{\mathbf{z}}{\sigma_G}\right\|^2\right],
\label{eq:score_matching_loss}
\end{equation}
where $\mathbf{s}_\theta$ is the score network parameterized by $\theta$, $\sigma_G$ is the noise level parameter analogous to the one used in KGMM, and the expectation is over data samples $\bm{\mu}$ and noise $\mathbf{z} \sim \mathcal{N}(0,\mathbf{I})$. We apply the same methodology as detailed in \citep{giorgini_response_theory} and its Appendix, but generalized to the higher-moment cases here. For example, at time $t=0$ the score function must satisfy the discrete identity $\langle g \mathbf{s} \rangle = -\langle \nabla g\rangle$ for any observable $g$.  This can serve as an aposteriori check on the fidelity of the score-matching procedure as well as a starting point for the a ``correction'' procedure as outlined in the Appendix of \citep{giorgini_response_theory}. An appropriately modified procedure was implemented for the results in the main text.

Denoising score-matching provides a scalable alternative to KGMM, enabling accurate score function estimation for high-dimensional systems without requiring dimensional reduction. The training procedure requires larger datasets and longer training times compared to KGMM, but provides the flexibility to handle arbitrary spatial perturbations in high-dimensional systems where traditional reduced-order modeling strategies are not applicable.

\subsection{KGMM, Denoising Score-Matching, and Architectures: Practical Guidance}
\label{subsec:kgmm_vs_unet}

When applying data-driven methods to estimate the score function there are two primary choices:
\begin{enumerate}
\item loss function,
\item neural network architecture.
\end{enumerate}
Other choices we do not discuss here include hyperparameter tuning, physics-informed constraints, optimization algorithms, software/hardware, arithmetic precision, and initialization (pre-trained/foundation vs.\ random).

By default, denoising score matching (DSM) and its variants~\citep{useful_diffusion} serve as a reliable choice, though they may use more computation than necessary. An alternative, KGMM, is a two-step procedure: (i) cluster the data to distill it to its most informative samples, then (ii) solve a standard regression to fit the score on those interpolation points. KGMM is most useful when cluster centroids lie on the attractor and the local effective dimension is modest (heuristically $ \leq \mathcal{O}(10)$). If local neighborhoods of the attractor are low-dimensional, clustering behaves well; if centroids drift off the manifold or too many clusters are required, use DSM instead. Dimensionality reduction techniques (PCA/EOFs or a shallow autoencoder) can help: if a small set of modes captures the statistics of interest, reduce the dimensionality first and then apply KGMM on the reduced coordinates. In regimes where clustering works, KGMM quickly reduces data volume and supplies stable interpolation points for the score. In our experiments on lower-dimensional systems, KGMM was often an order of magnitude faster than directly optimizing DSM. For high-dimensional fields such as two-dimensional Navier–Stokes, clustering offered little benefit.

For neural network architectures we used two baselines: a multilayer perceptron (MLP) and a U-Net. In practice, selecting an architecture is empirical and problem-specific; many rules of thumb have counterexamples.  Broadly, MLPs can be used as general-purpose interpolants for low-dimensional state representations (e.g., modal amplitudes). For higher-dimensional data with spatial or temporal coherence, convolutional encoders/decoders (U-Nets) consistently perform well by exploiting locality and weight sharing. Architecture selection is empirical; our choices reflect domain knowledge in both deep learning and the underlying physics.

\section{KGMM Performance with Reduced Data}\label{sm:kgmm_reduced}

To test the robustness of the KGMM algorithm, we repeated the analysis for the three reduced-order models using only $10^4$ uncorrelated samples, a hundred-fold reduction from the $10^6$ samples used in the main text. All algorithm hyperparameters, including the kernel width $\sigma_G$ and the number of clusters $N_C$, were kept identical to those in the main text.

\begin{figure}[H]
    \centering
    \includegraphics[width=\textwidth]{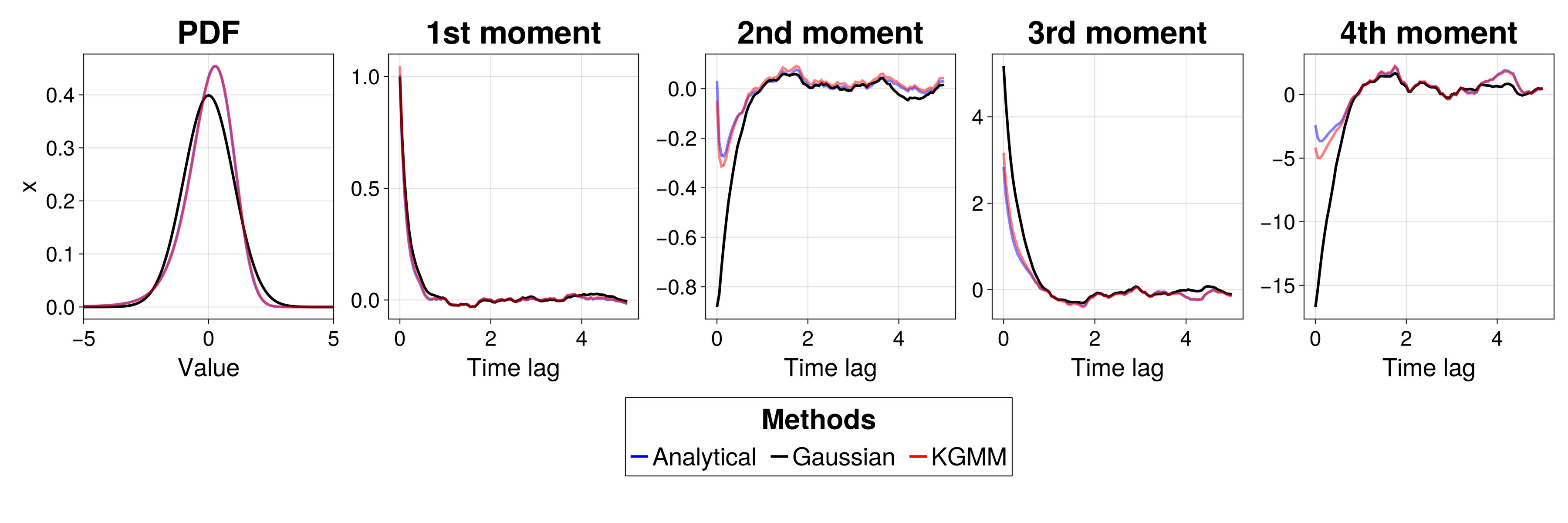}
    \caption{Same as Figure 1 in the main text, but with the KGMM score function estimated using only $10^4$ data points and unchanged hyperparameters.}
    \label{fig:sm_reduced_responses}
\end{figure}

\begin{figure}[H]
    \centering
    \includegraphics[width=\textwidth]{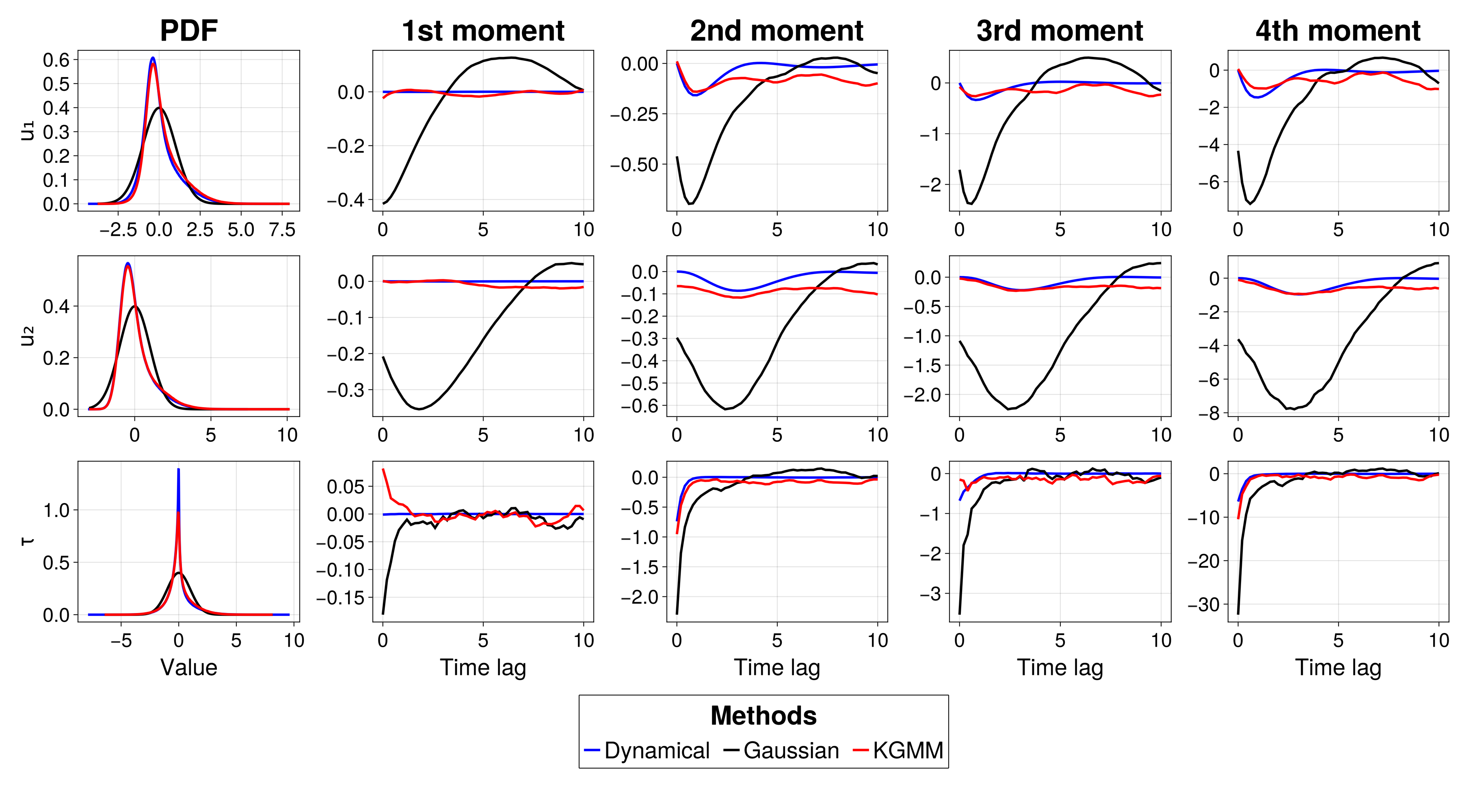}
    \caption{Same as Figure 2 in the main text, but with the KGMM score function for the slow-fast triad model estimated using only $10^4$ data points and unchanged hyperparameters.}
    \label{fig:sm_enso_responses}
\end{figure}

\begin{figure}[H]
    \centering
    \includegraphics[width=\textwidth]{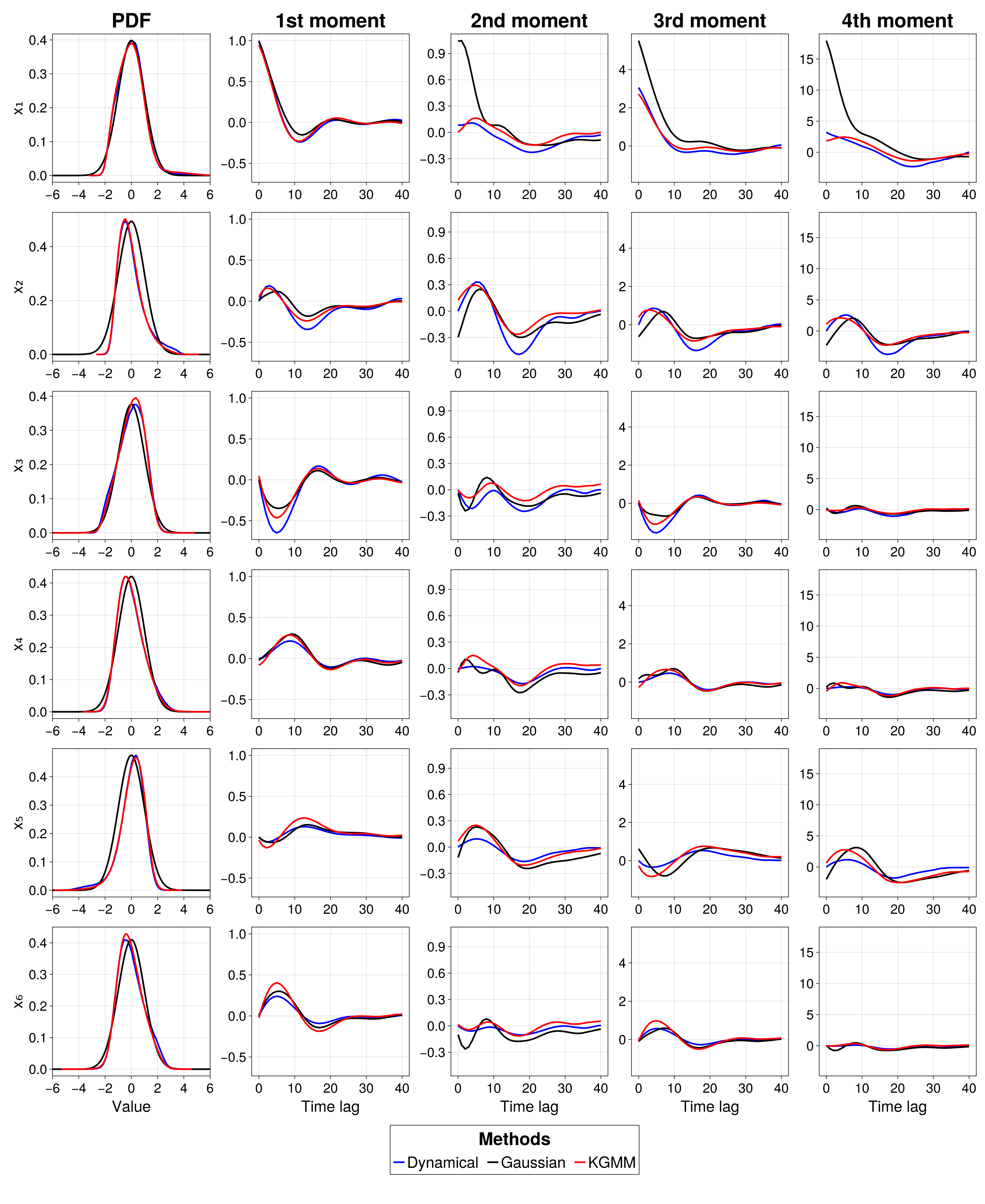}
    \caption{Same as Figure 3 in the main text, but with the KGMM score function for the stochastic barotropic model estimated using only $10^4$ data points and unchanged hyperparameters.}
    \label{fig:sm_cdv_responses}
\end{figure}

The results are shown in Figures \ref{fig:sm_reduced_responses}-\ref{fig:sm_cdv_responses}. Even with significantly less data and no re-tuning of parameters, the estimated response functions remain in excellent agreement with the ground truth and continue to substantially outperform the Gaussian approximation. This highlights the data efficiency of the KGMM framework.

